\documentclass[twoside]{article}

\usepackage{PRIMEarxiv}

\usepackage[utf8]{inputenc} % allow utf-8 input
\usepackage[T1]{fontenc}    % use 8-bit T1 fonts
\usepackage{hyperref}       % hyperlinks
\usepackage{url}            % simple URL typesetting
\usepackage{booktabs}       % professional-quality tables
\usepackage{amsfonts}       % blackboard math symbols
\usepackage{nicefrac}       % compact symbols for 1/2, etc.
\usepackage{microtype}      % microtypography
\usepackage{lipsum}
\usepackage{fancyhdr}       % header
\usepackage{graphicx}       % graphics
\graphicspath{{media/}}     % organize your images and other figures under media/ folder
\usepackage{subcaption}
\usepackage{placeins}
\usepackage{array}
\usepackage{enumitem}
\usepackage{hyperref}
\usepackage[numbers]{natbib}
\usepackage{doi}
\title{ModRes: A modification of the ResPlan dataset ready for floor plan analysis and graph machine learning}

\author{
  Emma Shelley\\
  School of Computational and Mathematical Sciences \\
  Cardiff University \\
  Cardiff\\
  \texttt{shelleye@cardiff.ac.uk} \\
   \And
  Wassim Jabi \\
  Welsh School of Architecture \\
  Cardiff University \\
  Cardiff\\
  \texttt{jabiw@cardiff.ac.uk} \\
   \And
  Thomas Woolley \\
  School of Computational and Mathematical Sciences \\
  Cardiff University \\
  Cardiff\\
  \texttt{woolleyt1@cardiff.ac.uk} \\
   \And
  Katerina Kaouri \\
  School of Computational and Mathematical Sciences \\
  Cardiff University \\
  Cardiff\\
  \texttt{kaourik@cardiff.ac.uk}}

\begin{document}
\maketitle

\begin{abstract}
Floor plan datasets underpin computational analysis of building layouts, spatial connectivity and graph machine learning. However, publicly available datasets often require substantial preprocessing before they can be used reliably. We present ModRes, a cleaned and enriched version of the ResPlan dataset, containing 8,431 valid and unique residential floor plans. Invalid and duplicate floor plans are removed through a dedicated cleaning workflow, producing a ready-to-use dataset. For each floor plan, ModRes provides two-dimensional geometry, images, three-dimensional building representations and three complementary dual graphs in CSV and JSON formats with graph-, node- and edge-level features. Continuous and categorical graph attributes support machine learning workflows. We characterise the dataset using graph-based statistics and demonstrate its utility through a range of case studies. ModRes provides structured geometric, topological and semantic representations needed to automate spatial analysis and support machine learning-based workflows for building design and performance evaluation.
\end{abstract}

% keywords can be removed

\section{Introduction}
Floor plans provide a geometric representation of an indoor environment, showing the structure of, and spatial relationships between, architectural spaces. Floor plans typically show features such as rooms, walls, and doors, and how these features are arranged and connected to each other. Floor plans are widely used across a range of fields, including architecture \citep{ahmed2011improved}, real estate, spatial analysis \citep{dao}, and robotics \citep{boniardi2019robot}. Floor plans provide a concise representation of a building's layout, making them valuable for design, analysis, and automated processing across these fields.

Spatial analysis techniques are widely used to evaluate how the layout of a building affect accessibility, visibility within the building and occupant behaviour \citep{tawfik, lu}. These techniques can be used to improve the functionality of a built environment. For example, when designing a shop, spatial analysis can be useful to ensure there is high visibility throughout the shop so that more products are seen by customers. 

An example of spatial analysis techniques is space syntax, in which floor plans are represented as a network that captures spatial relationships between different areas of a building, allowing researchers to evaluate how people may move through a building \citep{spacesyntax}. For example, space syntax could be utilised when designing an office building, to ensure meeting rooms are in an area of the building with high footfall. Furthermore, visibility within a building can be evaluated using isovist analysis, which measures the total area visible from different points in the building \citep{lonergan}. Isovist analysis helps to analyse the openness of a floor plan, which can be useful when designing a building where visiblity is a key design objective. For example, buildings where privacy and confidentiality are important, such as government buildings, would benefit from lower visibility values whereas buildings where collaboration and supervision are essential, such as an emergency department, require higher visibility values. 

More recently, due to advances in machine learning techniques, there is an increased interest in automated floor plan analysis \citep{pizarro}. Because of this interest, there is demand for large-scale floor plan datasets, which can be used for computational analysis, or training machine learning models. Floor plan datasets have been used for a range of applications, such as building and room classification \citep{paudel2021room} and accessibility analysis \citep{zhang}. Unlike other fields such as healthcare, the architecture, engineering and construction (AEC) sector lacks publicly available datasets, as floor plans are often not publicly released due to confidentiality risks. We address this lack of datasets in this paper, by presenting a preprocessed, publicly available dataset that is ready for use. This dataset is a cleaned and processed version of ResPlan \citep{resplan}.

\begin{figure}
    \centering
    \includegraphics[width=0.95\linewidth]{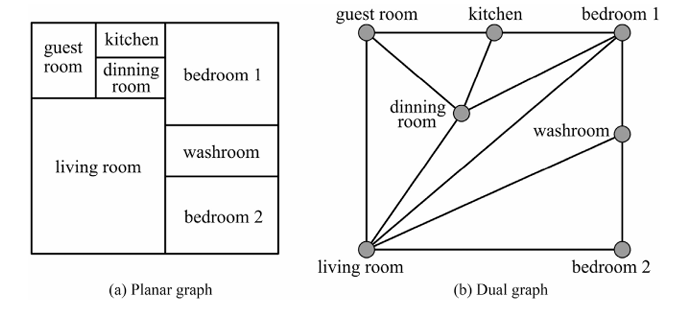}
    \caption{The planar (primary) and dual graph of a floor plan. Source: \cite{wang}.}
    \label{fig:dualgraph}
\end{figure}

Floor plan datasets can also be used in Graph Machine Learning (GML) \citep{hu}. GML is a branch of machine learning that works on structured data, where data points are represented by nodes, and their connections are shown by edges. Unlike traditional machine learning methods, GML methods extract relationships between different parts of the graph, instead of regarding them as separate entities. In the context of floor plans, a building can be represented as a graph. The primary (or planar) graph of a floor plan has edges that correspond to the walls of the building, and nodes representing the intersections between walls \citep{xie}. The dual graph, which we use in this work, has nodes representing architectural components such as rooms and doors, and edges representing relationships between these components (see Figure \ref{fig:dualgraph}) \citep{wang}.

Graph neural networks can then be trained for a range of tasks, such as building classification \citep{kong} or estimating energy usage \citep{jabi}. Representing floor plans as dual graphs allows machine learning methods to focus on the structure and connectivity of a building, instead of relying on the building's full geometry, which can be a more computationally efficient way of analysing buildings. Thus, we will incorporate multiple graphical representations of each floor plan into our dataset.
%explanation of resplan

\subsection{The ResPlan dataset}
ResPlan \citep{resplan} is a dataset of approximately 17,000 residential floor plans, developed from public real estate listings. Each floor plan is given in a fully vectorised format, with each component represented by a Shapely Polygon; a geometric feature representing a 2D area. Each Polygon is given semantic annotations denoting the type of geometry, e.g. `bedroom' or `wall'. There is also metadata attached to these floor plans, namely `unitType', which is a categorical variable describing the residence type (e.g. Apartment, or Villa), and `area', which is the full area of the house in square metres. 

ResPlan offers increased structural diversity compared to other floor plan datasets, with many rooms having non-rectangular shapes, and floor plans ranging from small one-bedroom apartments to large multi-wing homes.

However, in ResPlan, there are errors that need fixing, the details of which are outlined in the Section \ref{cleaning}. We remove these floor plans, as any results obtained using floor plans with errors will lack validity and applicability to real-life residences. Furthermore, there are many instances of duplicate floor plans within the dataset. This could have a large impact when training a machine learning model on the data, as the model will learn more instances of one floor plan than another, which can bias the results. Therefore, we also remove any duplicate floor plans.

\subsubsection*{Dual graphs of floor plans}
\begin{figure}
    \centering
    \includegraphics[width=0.75\columnwidth]{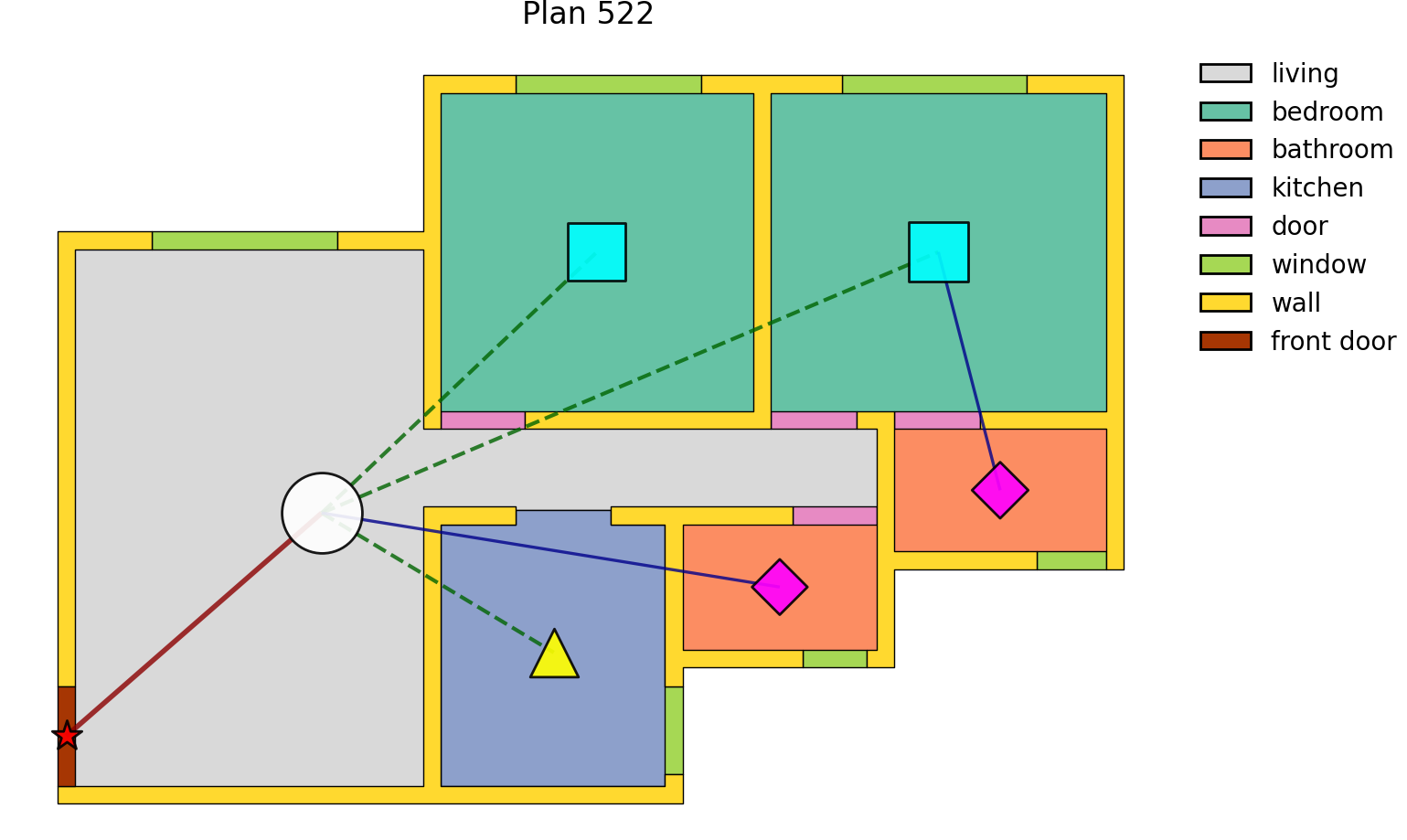}
    \caption{An example of a graph in ResPlan. Dark blue lines denote connections through doors, and dark green dashed lines represent connections through an arch. Source: \cite{resplan}.}
    \label{fig:resplangraph}
\end{figure}

Each floor plan in the ResPlan dataset is also converted into a room connectivity graph, where the nodes correspond to rooms, and the edges represent connections between these rooms. Rooms connected through a doorway are connected by an edge labelled `door', which is represented by a dark blue solid line (see Figure \ref{fig:resplangraph}). Rooms connected directly without a door are connected by an edge labelled `arch', and are represented by a dark green dashed line.

In Figure \ref{fig:resplangraph}, we observe that the edge labels are not consistent. Rooms that are connected by doorways, for example the two bedrooms and the living room, are connected by a green dashed line, which represents an `arch', meaning there is no door between them. However, we can clearly see that there are doors between them. These errors could lead to inaccuracies when using this dataset for spatial analysis, as the actual adjacency between rooms is often not represented within the graph. 

The graph inconsistencies are another key issue that we fixed in this paper. We create new graphical representations for each floor plan, with increased validity. To create these graphs, we convert each floor plan from a list of Shapely Polygons \citep{shapely_2024} to a list of TopologicPy Faces \citep{topologicpy}, and use TopologicPy's `Graph' class.

\subsection{TopologicPy}
TopologicPy is an open-source Python library designed for topological modelling and spatial analysis \citep{topologicpy}. TopologicPy provides data structures that can be used to represent architectural features, such as Vertices, Edges, Faces (2D shapes) and Cells (3D shapes). A key feature of TopologicPy is its ability to analyse spatial relationships between different architectural features, such as adjacency and connectivity. Unlike purely geometric representations, topological representations highlight the relationship between different spaces, making such representations useful when analysing indoor environments.

A core component of TopologicPy is its `Graph' class, in which geometric structures, such as a floor plan, can be converted into a graph that captures the spatial relationships between their components. This class also allows for computational analysis of graphs. Furthermore, TopologicPy also incorporates Graph Machine Learning, making it a useful way to represent a dataset, as it easily allows for future machine learning applications. 

\subsection{Aims and Paper outline}
We aim to create a publicly available dataset that is ready for use, developed from ResPlan.

This paper will be set out as follows:

First in Section \ref{datasets}, we look at previous floor plan datasets that are available, with a focus on the dual graphs of these floor plans. Then, Section \ref{cleaning} outlines the rigorous cleaning process we carry out on ResPlan to remove all invalid and duplicate floor plans. Section \ref{features} outlines each of the dual graphs in our dataset, followed by a description of the data availability and structure. Each of the graphs in the dataset are available in both CSV and JSON format. This broadens the scope of applications for this dataset, as it provides the user with a wide choice of data structures that can be used for different applications. This is followed by a discussion of the diversity of the dataset. Next, in Section \ref{energy} we show how ModRes can be used to carry out spatial analysis on a building, including energy simulations, which determine how much energy is needed to heat or cool the building. Finally, in Section \ref{lims}, we discuss the limitations of ModRes, and potential avenues for future improvements.

\section{Existing floor plan datasets}\label{datasets}
Floor plan datasets play a vital role in many applications in the architectural domain and beyond, including spatial analysis and machine learning algorithms \citep{rplan}. Some floor plan datasets have been developed, but they vary widely in quality, content, size, and structure, which can in turn affect their suitability for different applications. The representation of the data within these datasets is also important, as some datasets store floor plans as rasterised images \citep{rplan}, while others include graphs representing the adjacency between each feature of the floor plan \citep{msd}. We will now look at some widely used floor plan datasets, their strengths and their potential limitations for different applications.

\paragraph{Modified Swiss Dwellings (MSD)} \citep{msd} contains 5,300 multi-unit floor plans, made up of approximately 18,900 individual dwellings. This was developed from the Swiss Dwellings (SD) dataset, following a careful cleaning process, including removing any non-residential data points, and any near-duplicates. Each floor plan is an entire floor of a building, containing multiple apartments. This dataset is highly complex and diverse, making it a benchmark dataset for multi-unit residential buildings \citep{msd}. MSD also provides dual graphs for each floor plan, opening up its applications to both geometrical and relational analysis.

\paragraph{CubiCasa5k} \citep{cubi} is a dataset of approximately 5,000 floor plans from Finland, developed from real estate listings using a vectorisation pipeline. Each floor plan contains semantic annotations, meaning that every room and architecural feature is represented by a 2D shape, and assigned a semantic class (i.e.`kitchen', `window',`'bedroom', `door'). These annotations cover 80 categories, and were manually applied by a human, then checked by another human, providing a high level of accuracy. Despite these high-quality annotations, CubiCasa5k is relatively small compared to other available datasets. This limits the diversity of floor plans, restricting the applicability of machine learning models trained with this dataset.  

\paragraph{RPLAN} \citep{rplan} is among the largest publicly available floor plan datasets, consisting of approximately 80,000 floor plans from real residential buildings in Asia. Each floor plan is available in a rasterised format, consisting of a $256\times256$ pixel image, where each pixel has semantic annotations, denoting room types or structural components. While this representation is useful for image-based tasks, it can cause challenges when looking at the floor plan as a whole. For example, topological relationships between rooms cannot be extracted directly from the pixel data; the data must first be vectorised, converting regions of pixels into geometric shapes.

\begin{figure}
    \centering
    \begin{subfigure}[t]{0.49\columnwidth}
        \centering
        \includegraphics[width=\linewidth]{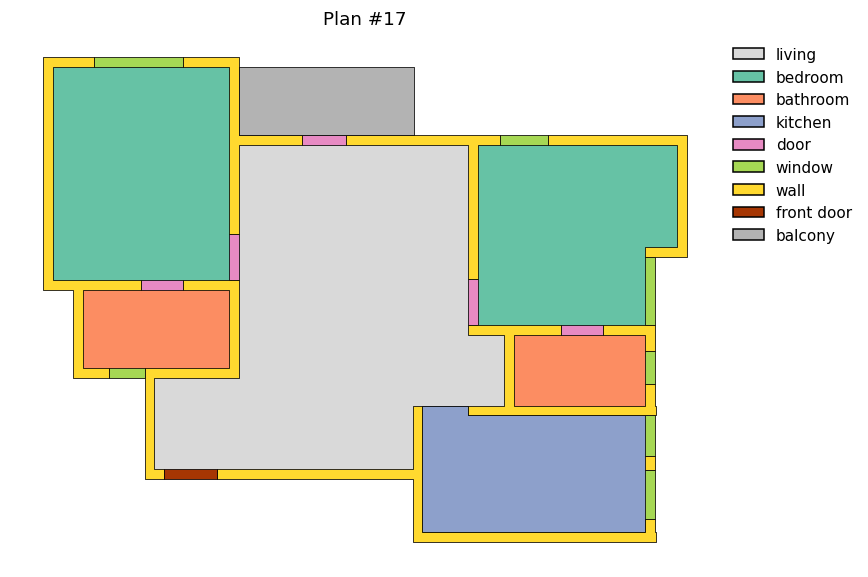}
        \caption{ResPlan. Source: \cite{resplan}.}
        \label{fig:datasetssub1}
    \end{subfigure}
    \begin{subfigure}[t]{0.49\columnwidth}
        \centering
        \includegraphics[width=\linewidth]{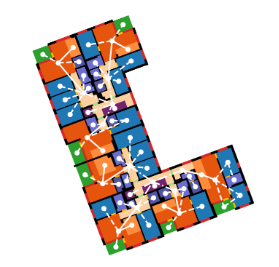}
        \caption{MSD. Source: \cite{msd}}
        \label{fig:datasetssub2}
    \end{subfigure}
    \vspace{0.5em}
    \begin{subfigure}[t]{0.49\columnwidth}
        \centering
        \includegraphics[width=\linewidth]{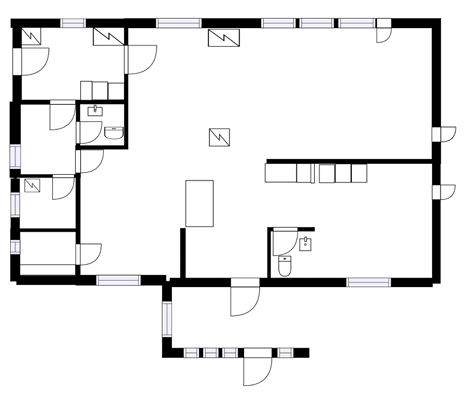}
        \caption{CubiCasa5k. Source: \cite{cubi}.}
        \label{fig:datasetssub1}
    \end{subfigure}
    \begin{subfigure}[t]{0.49\columnwidth}
        \centering
        \includegraphics[width=\linewidth]{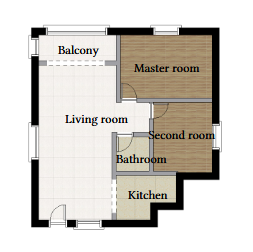}
        \caption{RPLAN. Source: \cite{rplan}.}
        \label{fig:datasetssub1}
    \end{subfigure}
    
    \caption{Examples of floor plans stored within different datasets.}
    \label{fig:datasets}
\end{figure}

\subsection{Dual graphs of floor plans}
While there is a range of floor plan datasets available, few of them represent these plans as graphs. Representing a floor plan graphically is beneficial, as the graph can show spatial relationships between different components, which can be used, for example, to assess connectivity of a building, or for graph neural networks (GNNs).

MSD provides two dual accessibility graphs for each of the floor plans \citep{msd}. One graph shows the connections between rooms, and the other graph shows the connections between `zones', i.e. whether a space is private, public, service, or outside. Each edge is classified into one of three groups depending on the connection type:
\begin{itemize}
    \item `passage' for when one can access the space without going through a door,
    \item `door' if spaces are connected by a door,
    \item `front door' if they are connected by a front door.
\end{itemize}
Nodes are also given semantic information, such as `room type'. While these graphs do contain a lot of information, they fail to include features such as windows or connections through walls, which can be important in applications such as energy analysis, as windows and walls affect how insulated a building is.

%which datasets represent them as graphs

ResPlan contains a dual graph for each floor plan \citep{resplan}. This graph shows the connectivity between rooms, with two rooms being connected by an edge if they are directly adjacent, either through a doorway or a wall. However, these graphs contain correctable errors, which we aim to fix in this work. This will be outlined in detail in Section \ref{cleaning}.

\section{Cleaning of ResPlan}\label{cleaning}
In this work, a Modified ResPlan dataset is generated from the ResPlan dataset. Below, we explicitly explain how the cleaning was enacted to ensure the validity and usability of each floor plan.

\subsection{Removal of invalid floor plans}
\begin{figure}[]
    \centering

    % Row 1
    \begin{subfigure}[t]{0.35\textwidth}
        \centering
        \includegraphics[width=\linewidth]{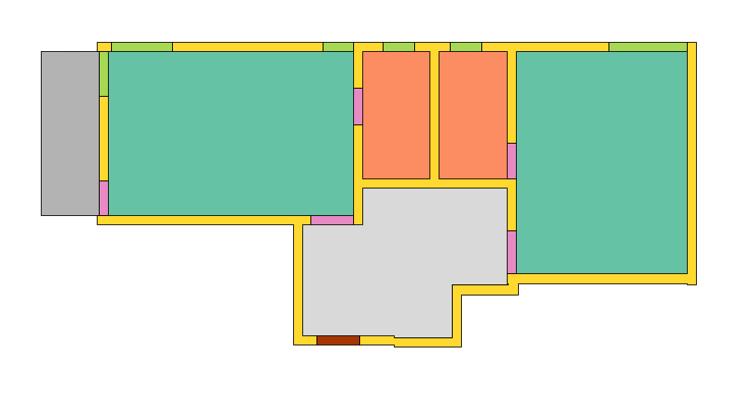}
        \caption{Essential rooms.}
        \label{fig:essentialroom}
    \end{subfigure}
    \hfill
    \begin{subfigure}[t]{0.35\textwidth}
        \centering
        \includegraphics[width=\linewidth]{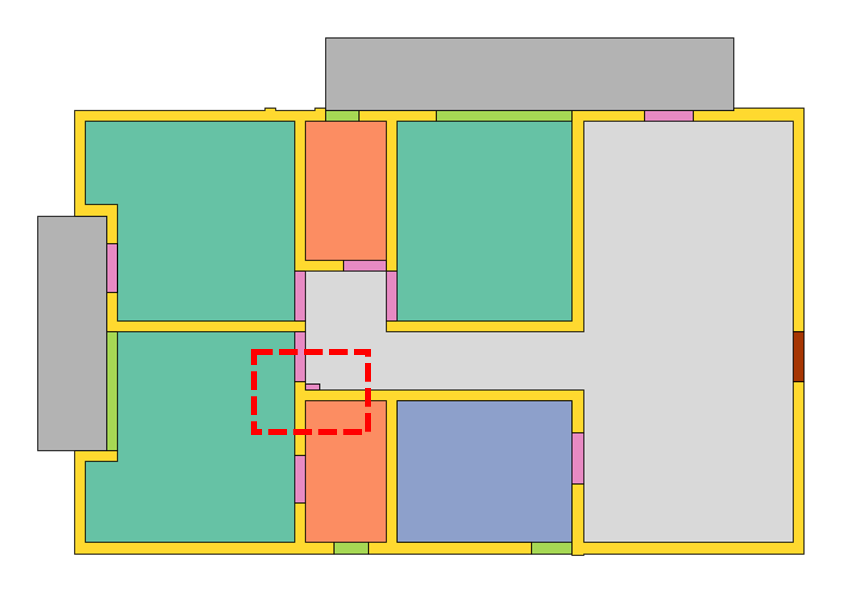}
        \caption{Invalid door.}
        \label{fig:invaliddoor}
    \end{subfigure}

    \vspace{0.2cm}

    % Row 2
    \begin{subfigure}[t]{0.35\textwidth}
        \centering
        \includegraphics[width=\linewidth]{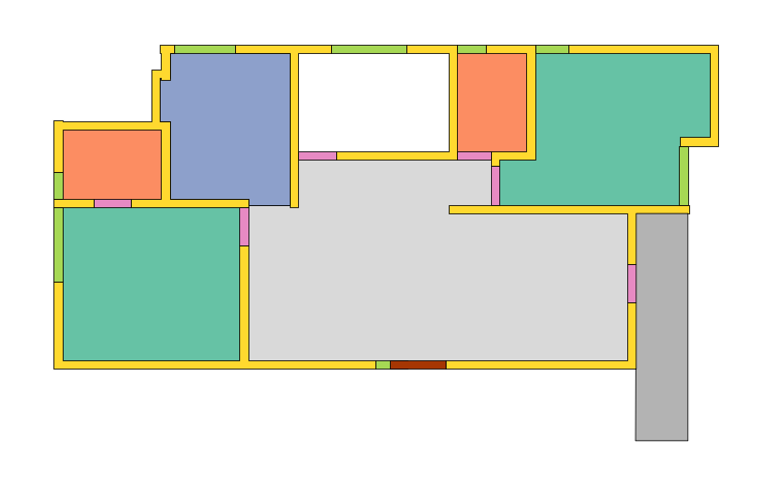}
        \caption{Missing room.}
        \label{fig:missingroom}
    \end{subfigure}
    \hfill
    \begin{subfigure}[t]{0.35\textwidth}
        \centering
        \includegraphics[width=\linewidth]{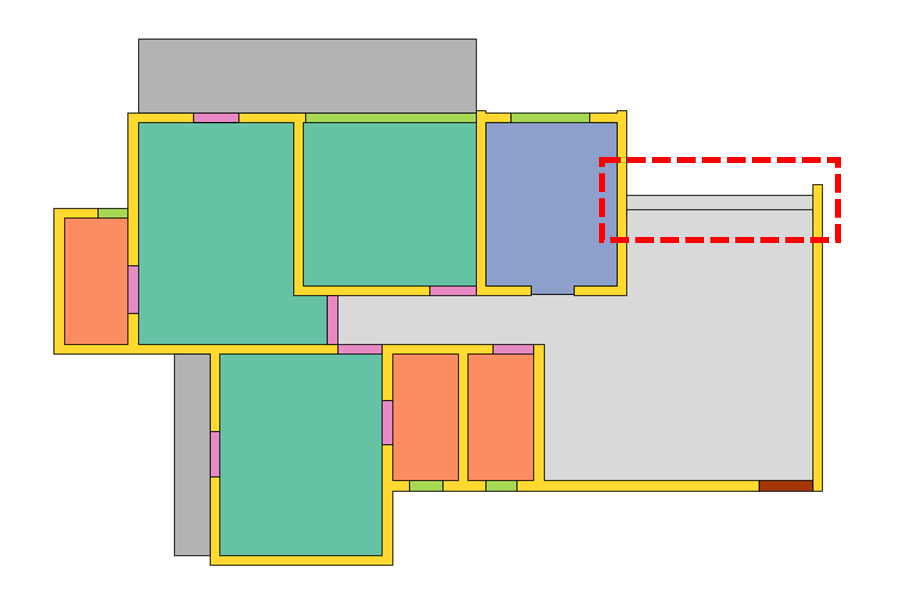}
        \caption{Missing wall.}
        \label{fig:missingwall}
    \end{subfigure}

    \vspace{0.2cm}

    % Row 3
    \begin{subfigure}[t]{0.35\textwidth}
        \centering
        \includegraphics[width=\linewidth]{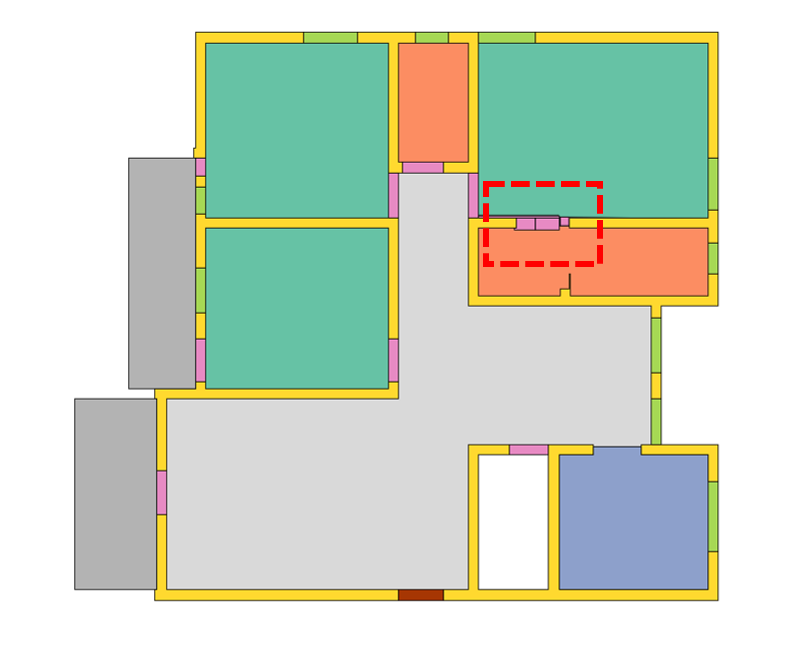}
        \caption{Door cluster.}
        \label{fig:doorcluster}
    \end{subfigure}
    \hfill
    \begin{subfigure}[t]{0.35\textwidth}
        \centering
        \includegraphics[width=\linewidth]{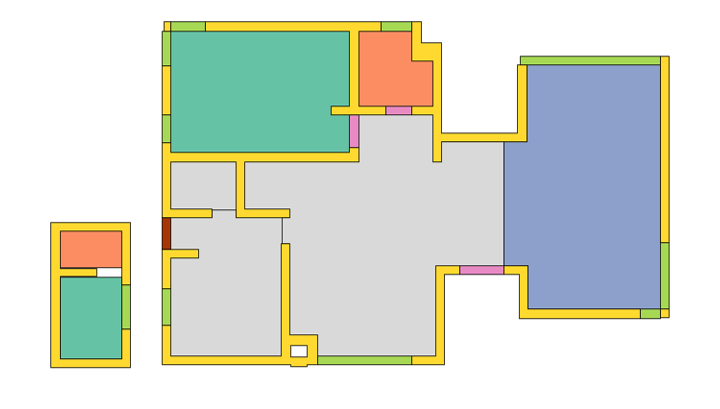}
        \caption{Disconnected graph.}
        \label{fig:disconnected}
    \end{subfigure}

    \vspace{0.2cm}

    % Row 4
    \begin{subfigure}[t]{0.35\textwidth}
        \centering
        \includegraphics[width=\linewidth]{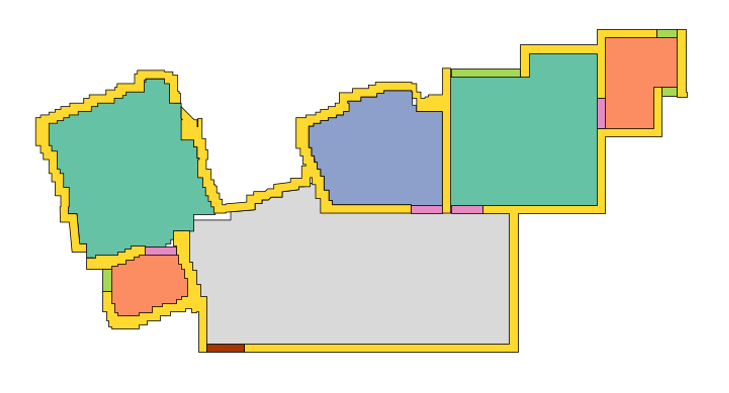}
        \caption{Invalid walls.}
        \label{fig:invalidwalls}
    \end{subfigure}

    \caption{Examples of invalid floor plans identified during the dataset filtering process.}
    \label{fig:invalid_examples}
\end{figure}

\begin{figure}
    \centering

    \begin{subfigure}[t]{0.49\columnwidth}
        \centering
        \includegraphics[width=\columnwidth]{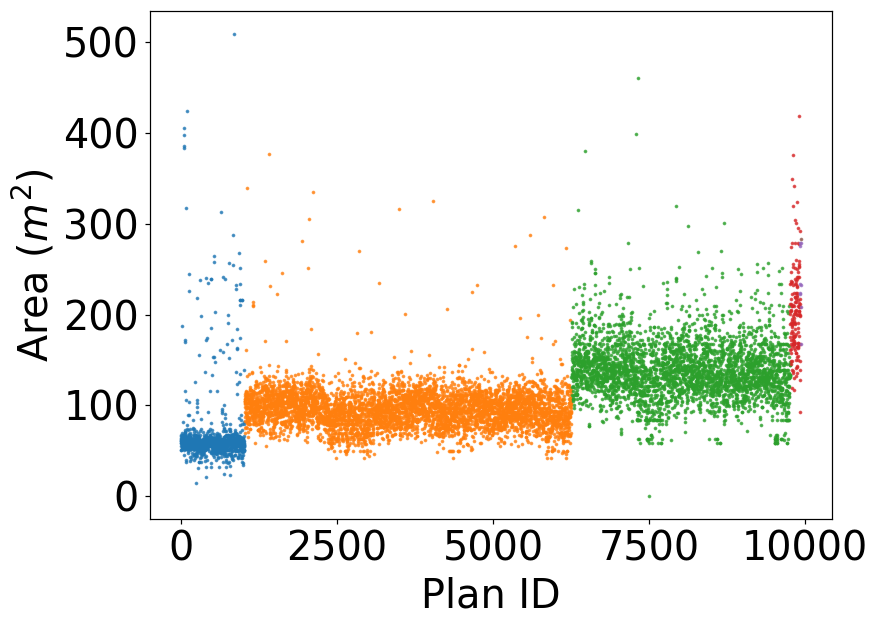}
        \caption{Distribution of floor plan areas before size filtering.}
        \label{fig:size1}
    \end{subfigure}
    \hfill
    \begin{subfigure}[t]{0.49\columnwidth}
        \centering
        \includegraphics[width=\columnwidth]{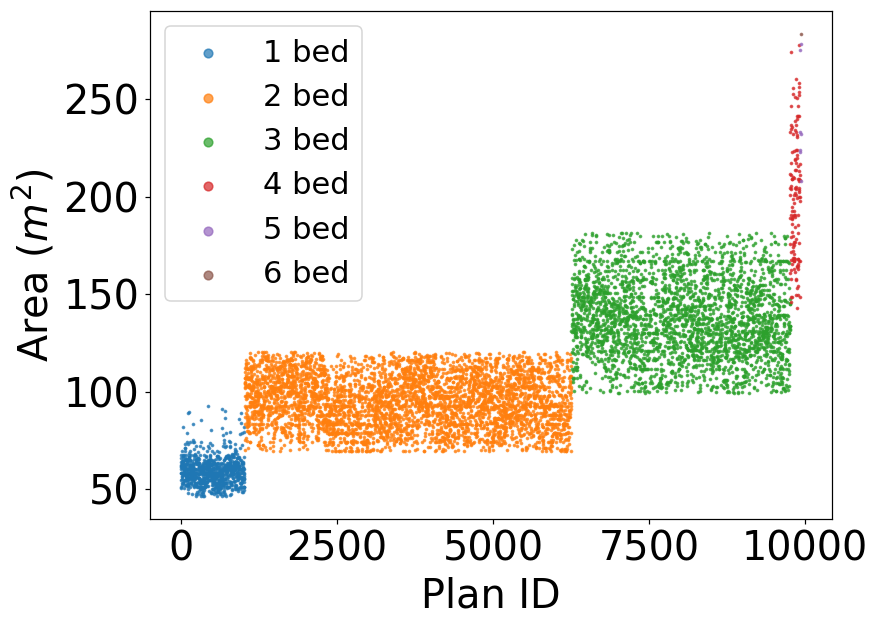}
        \caption{Distribution of floor plan areas after size filtering.}
        \label{fig:size2}
    \end{subfigure}

    \caption{The distribution of floor plan areas in the ResPlan dataset before and after applying the size filtering process.}
    \label{fig:size_filtering}
\end{figure}

First, we filtered through each floor plan to check its validity. We used the following steps:

\begin{itemize}
    \item \textbf{Essential rooms} Residential floor plans should contain the basic spaces required for occupancy. We remove plans without at least one living room, bedroom, bathroom, and kitchen (See Figure \ref{fig:essentialroom}).
    \item \textbf{Invalid doors} Plans with doors of unrealistically small dimensions are removed. We remove all doors under 4 units squared (ResPlan floor plans are not given to scale) (See Figure \ref{fig:invaliddoor}).
    \item \textbf{Missing rooms} Plans with large regions of missing geometry are removed. It is not possible for us to fill in each missing room as it is unknown what type of room they should be (See Figure \ref{fig:missingroom}).
    \item \textbf{Missing walls} Plans with gaps in the external envelope, leaving internal space exposed to the outside, are removed (See Figure \ref{fig:missingwall}). 
    \item \textbf{Door clusters} Plans with clustered door groups are removed. To identify them, each plan is first represented as an adjacency graph with room and door nodes. Any plan containing adjacent door nodes is excluded (See Figure \ref{fig:doorcluster}).
    \item \textbf{Disconnectedness} Plans are removed if they contain disjoint sections, or rooms with no accessible door connection. This is detected using the adjacency graph, excluding any plan with more than one connected component. We remove these plans as any house containing rooms that cannot be accessed by an occupant is unrealistic (See Figure \ref{fig:disconnected}).
    \item \textbf{Invalid walls} Plans with wall vertices clustered close together are removed. These clusters create degenerate edges when converting to other geometric representations (See Figure \ref{fig:invalidwalls}).
    \item \textbf{Size filtering} Plans with unrealistic recorded areas are removed. For each bedroom count, thresholds for area are specified in which we keep floor plans (See Table \ref{tab:filter}). These thresholds were calculated using a normal distribution, to find the $7.5$th and $92.5$th percentile. These values are consistent with real-life apartment sizes in the U.S. \citep{website}. Any plans outside these bounds are determined to be an outlier and discarded. See Figure \ref{fig:size1} and Figure \ref{fig:size2}.
\end{itemize}

\begin{table}
    \centering
    \begin{tabular}{|c|c|} \hline
        \textbf{Number of bedrooms} & \textbf{Area Threshold ($m^2$)}  \\
       \hline 1 & $[46.5, 92.7]$\\
        2 & $[69.7, 120.3]$\\
        3 & $[99.4, 181.4]$\\
        4 & $[142.6, 278.7]$\\
        5 & $[191.9, 278.7]$\\
         \hline
    \end{tabular}
    \caption{Thresholds for removal of floor plans in the size filtering step.}
    \label{tab:filter}
\end{table}

\subsection{Removal of duplicates}
The next step in the cleaning process was to remove any duplicates. This is essential so that the data is not biased towards particular plans. This could cause problems when training machine learning models, as the model will learn more from plans with more instances, which could skew the results.

For each plan, we made a vector representing its key values. These included the number of each room type, and the total area for each room type. This resulted in a vector with 7 values (number of kitchens is not included as all floor plans have one). We include: 
\begin{enumerate}
\item Number of bedrooms,
\item Total area of bedrooms,
\item Number of living rooms,
\item Total area of living rooms,
\item Number of bathrooms,
\item Total area of bathrooms,
\item Area of kitchen,
\end{enumerate}
These vectors were then compared with each other, and if any were identical, the second out of the two plans was removed. 

Table \ref{tab:removals} shows how many floor plans were removed at each step in the cleaning process. At the end of the cleaning process, we are left with 8431 unique and valid floor plans, approximately $49\%$ of the original dataset.

\begin{table}
    \centering
    \begin{tabular}{|p{2cm} |p{2cm} |p{2cm}|} \hline
        \textbf{Cleaning step} & \textbf{Number of plans removed} & \textbf{Percentage of plans removed} \\ \hline
        No kitchens & 84 & 0.5 \%\\
        Invalid doors & 627 & 3.7 \%\\
        Missing rooms & 1609 & 9.4 \%\\
        Missing walls & 372 & 2.2 \%\\
        Disconnected & 1606 & 9.4 \%\\
        Door Clusters & 2046 & 12.0 \% \\
        Size filtering & 1459 & 8.5 \% \\
        Duplicates & 833 & 4.9 \%\\
        Invalid walls & 40 & 0.2 \%\\ \hline
    \end{tabular}
    \caption{Number and percentage of floor plans in the ResPlan dataset removed through our cleaning process at each step.}
    \label{tab:removals}
\end{table}

\subsection{Geometry cleaning}
Finally, we clean up all geometry errors in the remaining floor plans in the dataset.

\begin{figure}
    \centering
    \includegraphics[width=\columnwidth]{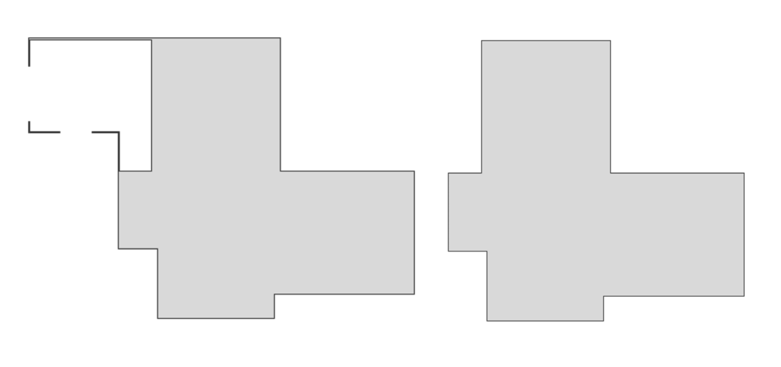}
    \caption{An example of small slivers being removed from the living room of plan 890 using the \texttt{`buffer\_shrink\_expand'} function.}
    \label{fig:slivers}
\end{figure}

\begin{figure}
    \centering
    \includegraphics[width=\columnwidth]{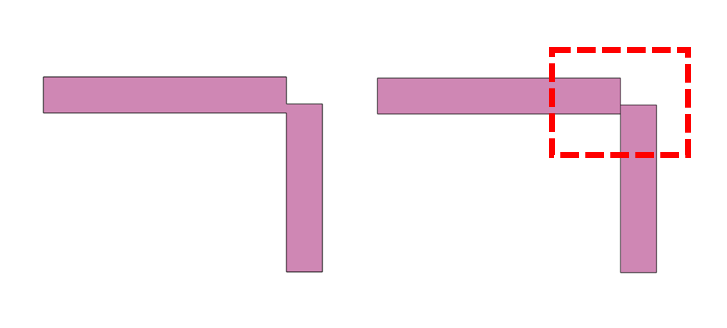}
    \caption{An example of a door geometry being separated into two distinct doors from plan 15633.}
    \label{fig:doors}
\end{figure}

\begin{itemize}
    \item \textbf{Scaling} In the original ResPlan dataset, none of the geometry was to scale. We used the reported area of each floor plan to scale them individually.
    \item \textbf{Buffer} We remove any small slivers from the geometry using ResPlan's \texttt{`buffer\_shrink\_expand'} function (see Figure \ref{fig:slivers}).
    \item \textbf{Door separation} Some floor plans include door geometries that represent more than one door. This causes problems when creating adjacency graphs of these floor plans, as two separate doors are represented by one node. Floor plans containing such geometries were identified by selecting each plan that contained a door with more than 4 vertices, implying it is non-rectangular. We then separate these geometries into two rectangles, representing two separate doors (See Figure \ref{fig:doors}).
    \item \textbf{Duplicate rooms} Some floor plans contain more than one geometry for the same room. We merge each of these duplicate geometries into one room.
\end{itemize}

\section{Features of the dataset} \label{features}
% converted to topologicpy
Originally, each ResPlan floor plan was stored as a single python dictionary, which is a collection of key-value pairs.  The keys of this dictionary are semantic categories (e.g. bedroom, bathroom, door, etc.) and the values were Shapely Polygon or MultiPolygon geometries.

We converted each of these Shapely Polygons into a TopologicPy Face by using a list of the vertices of the Polygon to construct a Face. Representing the geometry as a TopologicPy face allows us to easily represent spatial relationships between components. In particular, this representation enables identification of adjacency and connectivity between different elements in the floor plans, which proves particularly useful when constructing dual graphs of each floor plan.

\subsection{Graphs}
\begin{figure*}
    \centering
    \includegraphics[width=0.99\textwidth]{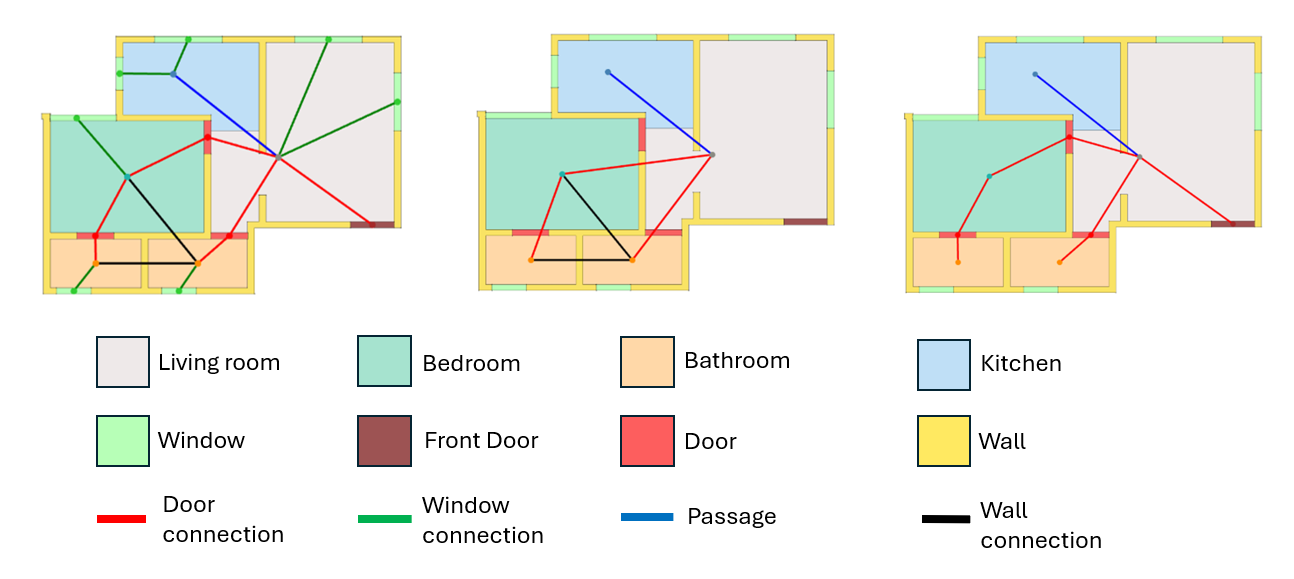}
    \caption{An example of each dual graph type for plan $\#1158$. From left to right they are the: Full graph, Adjacency Graph and Accessibility graph.}
    \label{fig:graphs}
\end{figure*}

For each floor plan, we have produced three graphs (see Figure \ref{fig:graphs}):
\begin{enumerate}
    \item \textbf{Accessibility Graph} shows the routes an individual could take when walking around the house. The graph nodes are doors and rooms, with edges that show the connections between rooms, either through direct connection or a doorway.
    \item \textbf{Adjacency Graph} only includes nodes for the rooms, and shows their connections to each other, either directly connected, through a door, or through a wall.
    \item \textbf{Full Graph} includes nodes for rooms, doors, and windows. The full graph shows how each room is connected to each other, like the adjacency graph, but also includes connections to doors and windows. 
\end{enumerate}

Each of these graphs has been enriched with data, which has been added to the dictionary of each node, edge, and graph. Data that were originally qualitative, e.g. room type, have been changed to be numerical, with a number assigned to each value instead of a word. For example, room type is stored as an integer value, 0-6, with each number corresponding to a different type of room.

%explain how we constructed these
\subsubsection{Graph}
\begin{figure}
    \centering
    \begin{subfigure}{0.3\columnwidth}
        \centering
        \includegraphics[width=\columnwidth]{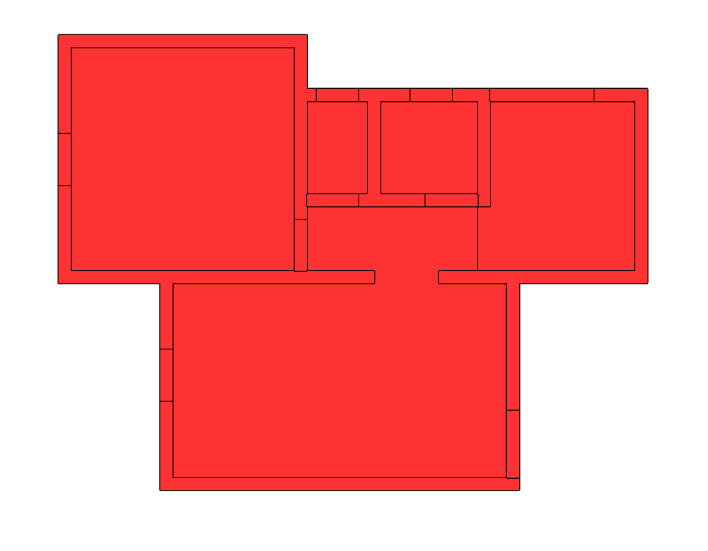}
        \caption{Area.}
        \label{fig:area}
    \end{subfigure}
    \begin{subfigure}{0.3\columnwidth}
        \centering
        \includegraphics[width=\columnwidth]{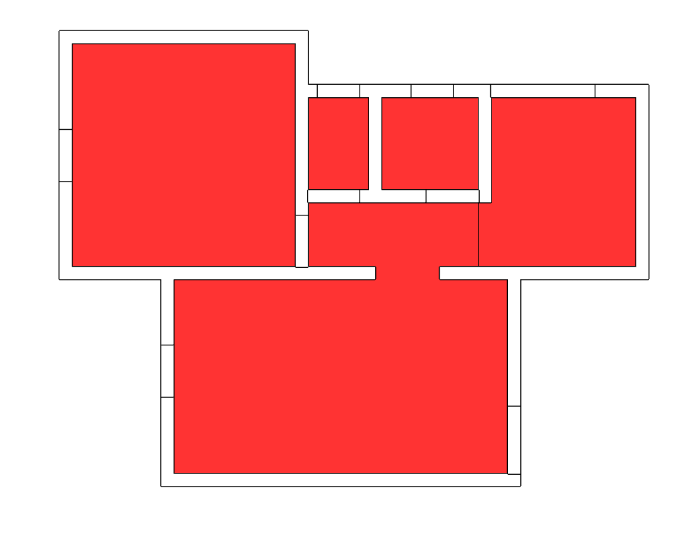}
        \caption{Net area.}
        \label{fig:netarea}
    \end{subfigure}
    \begin{subfigure}{0.3\columnwidth}
        \centering
        \includegraphics[width=\columnwidth]{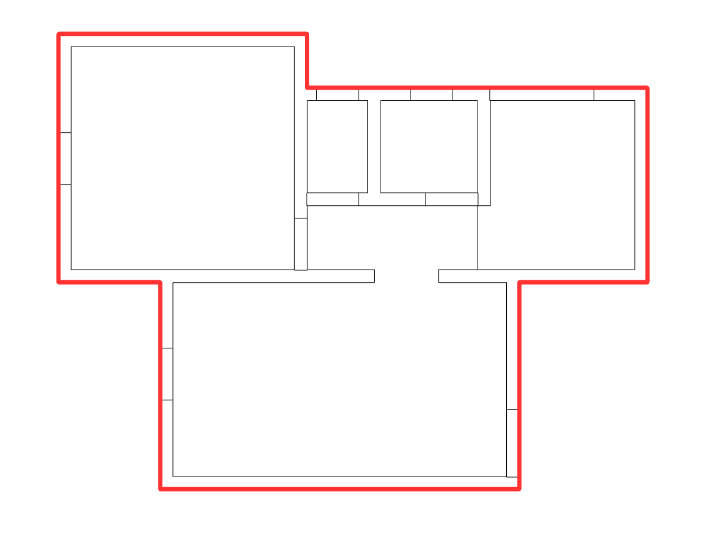}
        \caption{Perimeter.}
        \label{fig:perimeter}
    \end{subfigure}
    \caption{Diagrams showing how different graph features are calculated. Namely, the red areas or length of red line in (c). }
    \label{fig:nodes}
\end{figure}

\begin{figure}

    \centering
    
    % Top row
    \begin{subfigure}[t]{0.49\columnwidth}
        \centering
        \includegraphics[width=\columnwidth]{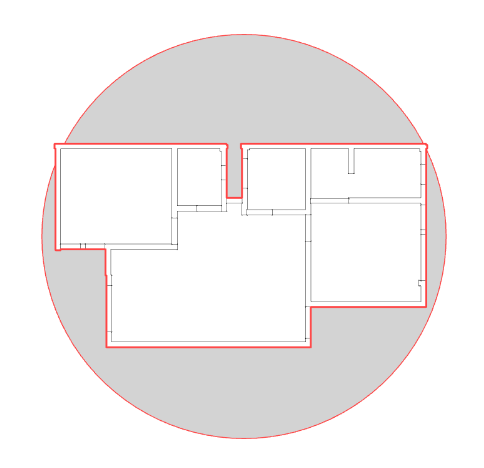}
        \caption{Compactness = 0.23.}
        \label{fig:low}
    \end{subfigure}
    \hfill
    \begin{subfigure}[t]{0.49\columnwidth}
        \centering
        \includegraphics[width=\columnwidth]{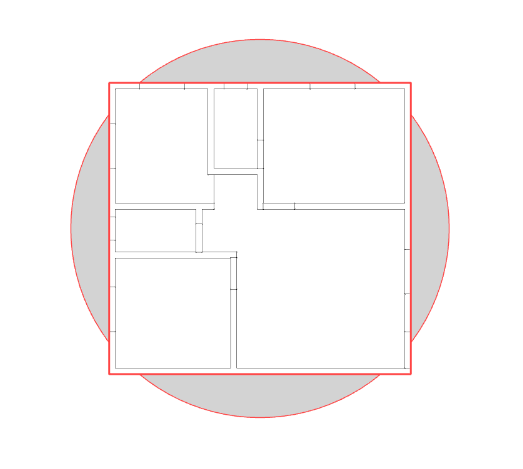}
        \caption{Compactness = 0.79.}
        \label{fig:high}
    \end{subfigure}

\caption{An example of high and low compactness floor plans, as calculated using the isoperimetric quotient. }
\label{fig:compactness}
\end{figure}

We add the following data to the dictionary for the whole graph:
\begin{enumerate}[label=\roman*.]
    \item \textbf{Number of bedrooms (minus one).} We record how many bedrooms the floor plan has, but, for consistency, as all other dictionary values begin at 0, we begin our indexing here at 0. Thus, the recorded value is one less than the number of bedrooms.
    \item \textbf{Area, $A$.} The total area of the plan, including walls, doors, and windows (see Figure \ref{fig:area}).
    \item \textbf{Net Area.} The sum of the area of each room in the floor plan, i.e. the total usable space (see Figure \ref{fig:netarea}).
    \item \textbf{Perimeter, $L$.} The perimeter of the entire floor plan (see Figure \ref{fig:perimeter}.
    \item \textbf{Compactness, $C$.} We calculate compactness using the isoperimetric quotient,
    $$C=\frac{4\pi A}{L^2},$$
    where $A$ is the area of the floor plan and $L$ is the perimeter, and this gives a value between 0 and 1 \citep{chakerian}. The quotient is the ratio between the area of the floor plan and the area of a circle with the same perimeter as the floor plan. This measurement is useful for specific applications, such as energy analysis, as the more compact a house is, the more energy efficient it is \citep{pomponi}. See Figure \ref{fig:compactness} for examples.
    
    \item \textbf{Unit type.} In ResPlan, there are four labels which describe the type of residential unit that each floor plan is. These are `Apartment', `IndependentHouse', `Villa' and `BuilderFloor'. We use one-hot encoding to store these within the dictionary of the graph, under the keys \texttt{'unit\_type\_0'} to \texttt{'unit\_type\_3'} respectively. One hot encoding converts categorical data into a form that machine learning can understand. Each category has a binary value, with 1 for the presence of the category, and 0 elsewhere. This prevents a machine learning model from learning an order from categories when there is not an order.
\end{enumerate}

\begin{table*}
    \centering
    \begin{tabular}{|c|c|c|c|c|c|} \hline
        \textbf{Room type} & \textbf{Measurement} & \textbf{Small} & \textbf{Medium} & \textbf{Large} & \textbf{Extra Large} \\ \hline
        Living room & Area & $(0, 20]$ & $(20, 40]$ & $(40, 60]$ & $(60, \infty)$ \\
         & Perimeter & $(0, 20]$ & $(20, 35]$ & $(35, 50]$ & $(50, \infty)$\\ \hline
        Bedroom & Area & $(0, 15]$ & $(15, 20]$ & $(20, 25]$ & $(25, \infty)$\\
         & Perimeter & $(0, 15]$ & $(15, 20]$ & $(20, 25]$ & $(25, \infty)$\\ \hline
        Bathroom & Area & $(0, 2.5]$ & $(2.5, 5]$ & $(5, 7.5]$ &$(7.5, \infty)$ \\
         & Perimeter & $(0, 6]$ & $(6, 9]$ & $(9, 12]$ & $(12, \infty)$\\ \hline
        Kitchen & Area & $(0, 5]$ & $(5, 10]$ & $(10, 15]$ & $(15, \infty)$\\
         & Perimeter & $(0, 10]$ & $(10, 13]$ & $(13, 16]$ & $(16, \infty)$\\ \hline
        Window & Area & $(0, 1]$ & $(1, 2]$ & $(2, 3]$ & $(3, \infty)$ \\
          & Perimeter & $(0, 4]$ & $(4, 6]$ & $(6, 8]$ & $(8, \infty)$ \\ \hline
        Door & Area & $(0, 1]$ & $(1, 2]$ & $(2, 3]$ & $(3, \infty)$ \\
          & Perimeter & $(0, 5]$ & $(5, 6]$ & $(6, 7]$ & $(7, \infty)$ \\ \hline
        Plan & Area & $(0, 100]$ & $(100, 150]$ & $(150, 200]$ & $(200, \infty)$ \\
          & Perimeter & $(0, 40]$ & $(40, 55]$ & $(55, 70]$ & $(70, \infty)$ \\ \hline
    \end{tabular}
    \caption{Measurement thresholds for each category for area and perimeter in the categorical dataset. Areas are in $m^2$ and perimeters are in $m$. }
    \label{tab:thresholds}
\end{table*}

\subsubsection{Nodes}
The nodes of the graph are located on the centroid of each geometry. We add the following data to the dictionary of each node:
%list of all different data saved in nodes
\begin{enumerate}[label=\roman*.]
    \item \textbf{Room type.} There are seven room types, stored numerically as 0-6. These correspond to 0 - living room, 1 - bedroom, 2 - bathroom, 3 - kitchen, 4 - window, 5 - door, and 6 - front door, respectively.
    \item \textbf{Area.} For rooms, the area is simply the area of the room. For doors and windows, the area is the area of the door or the window. To calculate this, we find the length of the door or window from the geometry, and then multiply it by its height. We assume a standard height of 2 metres for doors, and 1.2 metres for windows (see Figure \ref{fig:area}).
    \item \textbf{Perimeter.} As with the area, for rooms, the perimeter is simply the perimeter of the room. For doors and windows it is the perimeter of the door or window, respectively, calculated using the same standard heights stated in (ii) above. (see Figure \ref{fig:perimeter}).
    \item \textbf{Compactness.} We calculate the compactness of each room using the isoperimetric quotient, as we do for each floor plan.
\end{enumerate}
%explain difference between categorical and continuous csv file
\begin{figure}
    \centering
    % Row 1
    \begin{subfigure}[t]{0.49\columnwidth}
        \centering
        \includegraphics[width=\columnwidth]{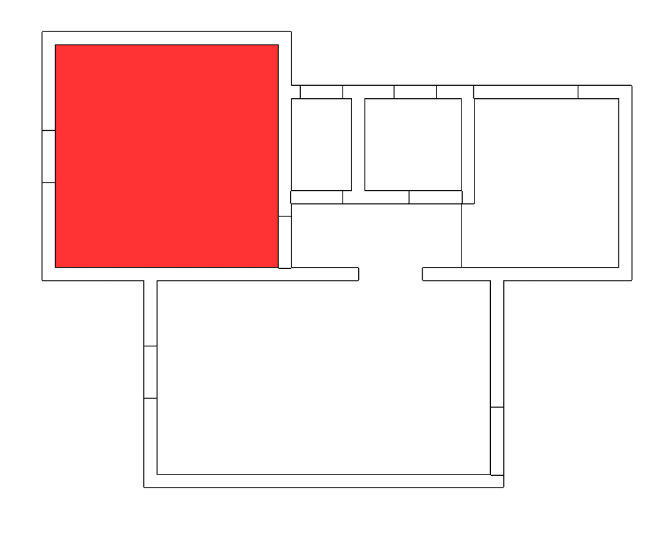}
        \caption{Room area.}
        \label{fig:roomarea}
    \end{subfigure}
    \begin{subfigure}[t]{0.49\columnwidth}
        \centering
        \includegraphics[width=\columnwidth]{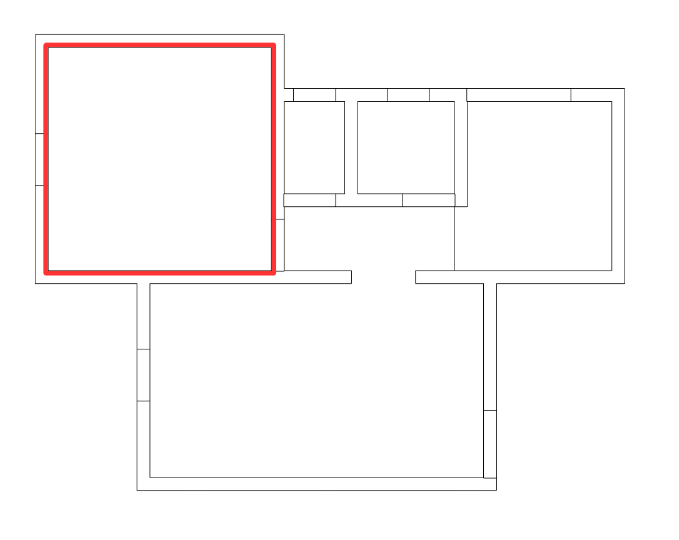}
        \caption{Room perimeter.}
        \label{fig:roomperim}
    \end{subfigure}
    % Row 2
    \begin{subfigure}[t]{0.49\columnwidth}
        \centering
        \includegraphics[width=\columnwidth]{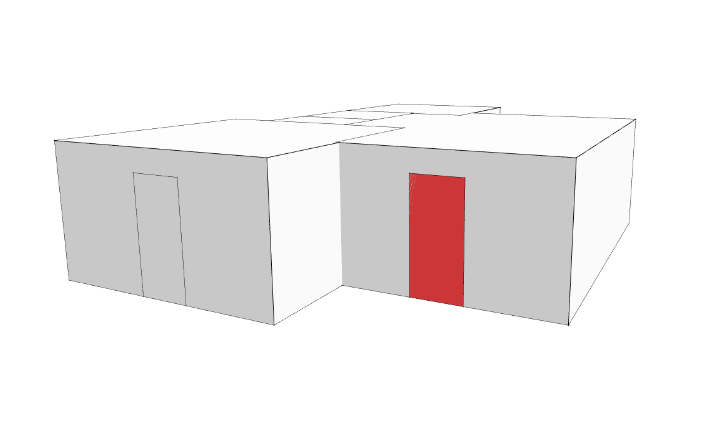}
        \caption{Door area.}
        \label{fig:doorarea}
    \end{subfigure}
    \begin{subfigure}[t]{0.49\columnwidth}
        \centering
        \includegraphics[width=\columnwidth]{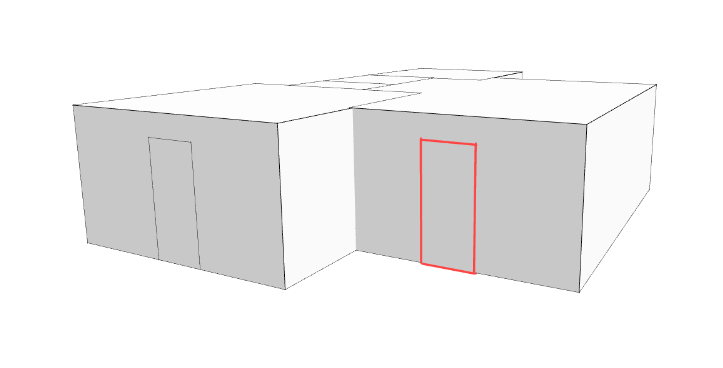}
        \caption{Door perimeter.}
        \label{fig:doorperim}
    \end{subfigure}
    \begin{subfigure}[t]{0.49\columnwidth}
        \centering
        \includegraphics[width=\columnwidth]{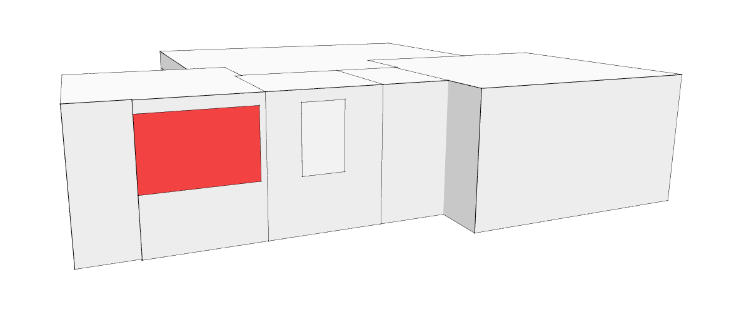}
        \caption{Window area.}
        \label{fig:windowarea}
    \end{subfigure}
    \hfill
    \begin{subfigure}[t]{0.49\columnwidth}
        \centering
        \includegraphics[width=\columnwidth]{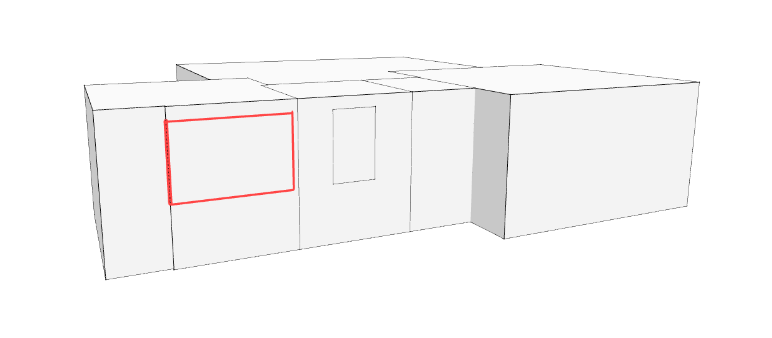}
        \caption{Window perimeter.}
        \label{fig:windowperim}
    \end{subfigure}

    \caption{Diagrams showing how different node features are calculated depending on the node type.}
    \label{fig:six}
\end{figure}

\subsubsection{Edges}
In the graphs, we have a total of four types of edges (numbered 0-3). The edge value is saved in the dictionary for the edge, under the key \texttt{edge\_type}. 0 is a `doorway', used either when the edge connects to a door, or in the case of the adjacency graph, when two rooms are connected by a door; 1 is a `passage', when two rooms are directly connected without a door; 2 is a `wall', when two rooms share a wall, but are not directly connected; and 3 is a `window', used only in the full graph, which is a connection between a room and a window.

\subsection{File format}

\begin{table*}[t]
\centering
\caption{Attributes included in the dataset.}
\label{tab:attributes}
\small
\begin{tabular}{p{2cm} p{1cm} p{1.8cm} p{1cm} p{1cm} p{1.5cm}  p{2.8cm} p{1.2cm}}
\hline
Attribute & Entity Type & Data Type & Units & Allowed Range & Encoding Method  & Source Method & Available at Inference Time \\
\hline

Bedrooms & Graph & Ordinal & -- & $0$--$6$ & Integer encoding & Computed from geometry & Yes \\
Area & Graph & Continuous & $\mathrm{m}^2$ & $>0$ & -- &  Computed from geometry & Yes \\
Net Area & Graph & Continuous & $\mathrm{m}^2$ & $>0$ & --  & Computed from geometry & Yes \\
Perimeter & Graph & Continuous & $\mathrm{m}$ & $>0$ & --  & Computed from geometry & Yes \\
Compactness & Graph & Continuous & - & $0$--$1$ & -- & Isoperimetric quotient & Yes \\
Unit Type & Graph & Categorical & -- & $0$--$3$ & One-hot  & Given in ResPlan & Yes \\

\hline

Room Type & Node & Categorical & -- & $0$--$6$ & Integer encoding &  Given in ResPlan & Yes \\
Area & Node & Continuous & $\mathrm{m}^2$ & $>0$ & -- & Computed from geometry & Yes \\
Perimeter & Node & Continuous & $\mathrm{m}$ & $>0$ & -- &  Computed from geometry & Yes \\
Compactness & Node & Continuous & - & $0$--$1$ &  -- & Isoperimetric quotient & Yes \\

\hline

Connection Type & Edge & Categorical & -- & $0$--$3$ & Integer encoding &  -- & Yes \\

\hline
\end{tabular}
\label{tab:schema}
\end{table*}

\subsubsection{Geometry}
\paragraph{Two-dimensional}
We generate JSON files for each floor plan that consist of just their geometry. This is stored as a dictionary, with keys corresponding to semantic categories, e.g. `bedroom', or `wall', and values that consist of a list of geometries that belong in that category. These geometries are stored as lists of coordinates, allowing a high level of flexibility. The list of coordinates allows for easy conversion to different geometric representations, for example Shapely geometries, or TopologicPy Faces, depending on how they will be used in applications. 

\paragraph{Three-dimensional}
We provide 3D versions of each floor plan. However, since ResPlan does not include any 3D data on the floor plans, such as the height of the ceilings or the shape of the roof, there are some assumptions made. Firstly, the International Residential Code states a minimum height of 7 feet for ceilings \citep{icc}, so we assume a height of approximately 8 feet, which is 2.4 m. For doors, we assume a height of 2 m, and for windows, a height of 1.2 m, positioned 1 m above the ground. Certain 3D analysis workflows, such as energy simulations using EnergyPlus \citep{energyplus}, require zero-thickness walls. However, ResPlan provides each floor plan with a uniform wall thickness, so we offset each of the walls. We offset internal walls by half the wall thickness to ensure they meet in their midline, and external walls by the full wall thickness. 

Each of the 3D floor plans is given as two BREP files. One BREP file contains the CellComplex of the building, which is a collection of TopologicPy Cells, while the other BREP file contains the apertures, i.e. the windows and external doors. 
\begin{figure}
    \centering
    \begin{subfigure}[t]{0.49\columnwidth}
        \centering
        \includegraphics[width=\columnwidth]{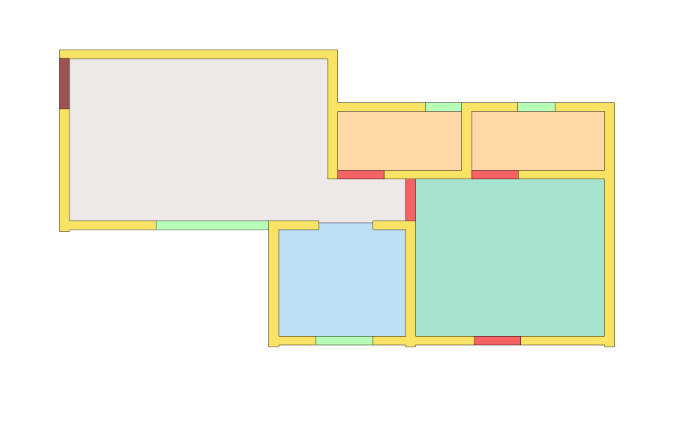}
        \caption{Two-dimensional}
        \label{fig:2d}
    \end{subfigure}
    \hfill
    \begin{subfigure}[t]{0.49\columnwidth}
        \centering
        \includegraphics[width=\columnwidth]{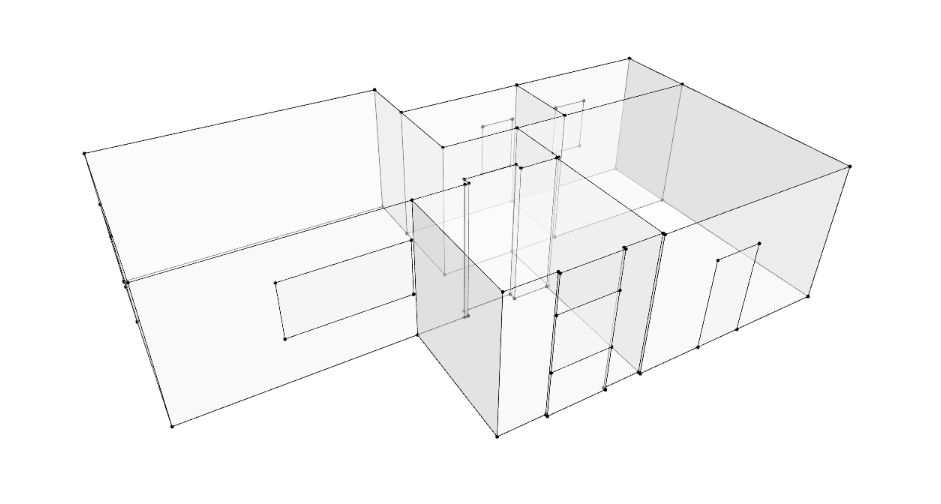}
        \caption{Three-dimensional}
        \label{fig:3d}
    \end{subfigure}
    \caption{Floor plan 67 in two-dimensional and three-dimensional form.}
    \label{fig:dimension}
\end{figure}

\begin{figure}
    \centering

    \begin{subfigure}[t]{0.49\columnwidth}
        \centering
        \includegraphics[width=\linewidth]{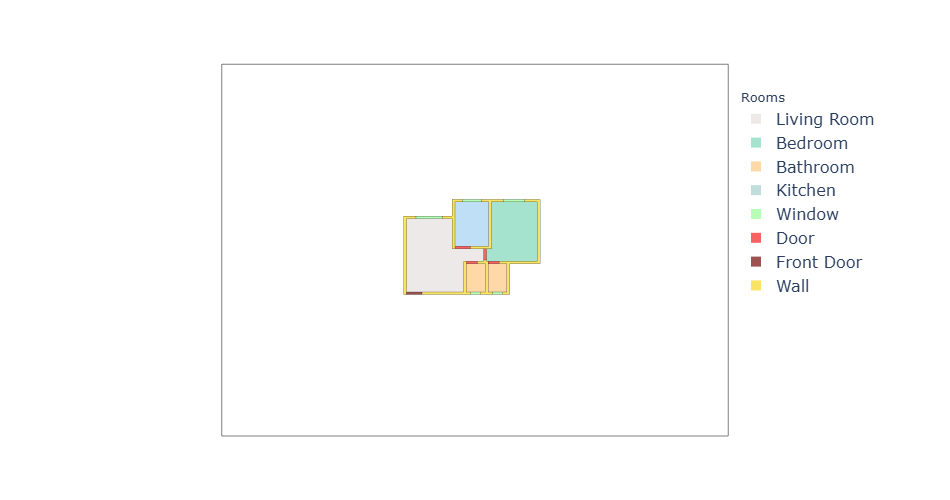}
        \caption{Plan 118.}
        \label{fig:image1}
    \end{subfigure}
    \hfill
    \begin{subfigure}[t]{0.49\columnwidth}
        \centering
        \includegraphics[width=\linewidth]{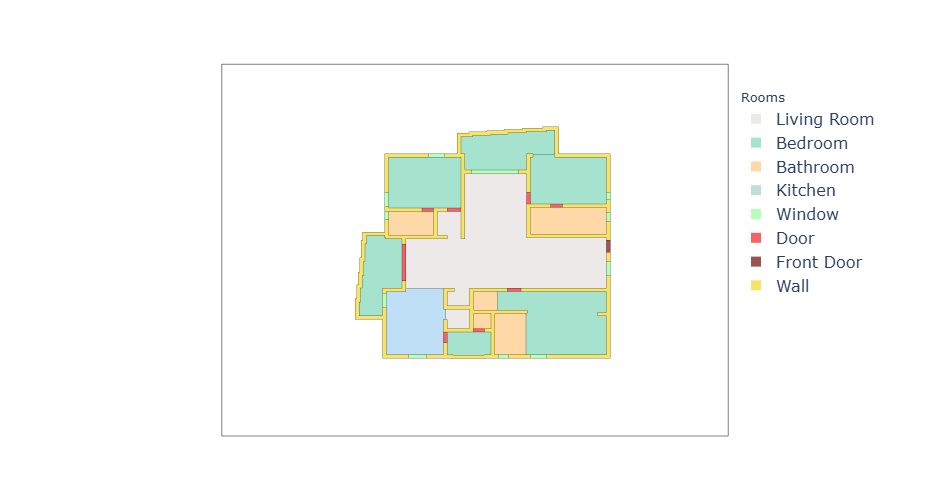}
        \caption{Plan 8430.}
        \label{fig:image2}
    \end{subfigure}

    \caption{Examples of the geometry images stored for each floor plan.}
    \label{fig:images}
\end{figure}

\paragraph{Images}
We also provide images of each of the floor plans. These are to scale, and are all contained within a bounding box of 35 metres wide and 30 metres long. This allows easy visual comparison between different sizes and layouts. See Figure \ref{fig:images} for examples. 

\subsubsection{Graph files}

\paragraph{Continuous data vs categorical data}
Measurements, such as area and perimeter of each of the rooms, are calculated to two decimal places. However, in practical settings, obtaining measurements to this level of accuracy is often unrealistic. Therefore, we generate another version of this dataset, where the features of each node and graph are not continuous but categorical. 

In this representation, each room is given two values, a category for the area (size), and a category for the perimeter (perimeter). Both categories take integer values, between 0 and 3, corresponding to small, medium, large, and extra large. 

The thresholds for these categories were determined separately for each room type, reflecting the fact that typical measurements vary widely between room types. For instance, a `large' bathroom is likely to be much smaller than a `large' living room (see Table \ref{tab:thresholds}). 

\paragraph{CSV}
For each type of graph, we provide a set of CSV files. The set consists of three files: \texttt{graphs.csv}, \texttt{nodes.csv}, and \texttt{edges.csv}, containing information on all the floor plans in the dataset.

\texttt{graphs.csv} contains a row of information for each graph, with the graph id identifying the corresponding floor plan in the dataset. The label for each graph is the number of bedrooms. However, for consistency we begin our indexing at 0, so the label is equal to one less than the number of bedrooms. There are then a number of columns for graph-level features. These contain information from the dictionary of each graph, although their form differs depending on whether the data is continuous or categorical (see Table \ref{tab:schema}).

\texttt{nodes.csv} contains information for each node across all graphs. Each node has a graph id, which identifies which graph it is from, and then a node id, uniquely identifying the node within that graph. The node has a label, which is an integer 0-6 corresponding to the room's type, as described previously. In addition, there are node-level features in the subsequent columns. As with the graph-level features, these differ depending on whether the data are categorical or continuous. 

\texttt{edges.csv} contains information on each of the edges of each graph. As with the nodes, graph id identifies which graph the edge is from. There is then a source id (\texttt{src\_id}) and a destination id (\texttt{dst\_id}), which tell us which nodes of the graph the edge is connected to. As these graphs are undirected, it contains two rows of data for each graph, one for each direction of the edge. The edge label (an integer 0 to 3 inclusive) corresponds to the connection type, as specified in the Graphs section above.

\paragraph{JSON}
We also generate a JSON file for every graph in the dataset. This consists of dictionaries for the whole graph, each node, and each edge. The dictionaries contain all the data outlined in the section above, as well as the geometry of each room, and colours that we use when plotting each of the graphs, ensuring reproducibility.

\paragraph{Images}
For each graph, we generate an image in the form of a PNG file. These images are the same structure as those in the geometry section, but they also have the graph included in the image. 

\subsection{Dataset diversity statistics}
For a dataset to be widely applicable, it must include a diverse set of data points. In this section, we assess the diversity of the Modified ResPlan dataset we have generated using graph-based metrics.

\begin{figure}
    \centering
    \begin{subfigure}{0.49\columnwidth}
        \centering
        \includegraphics[width=\columnwidth]{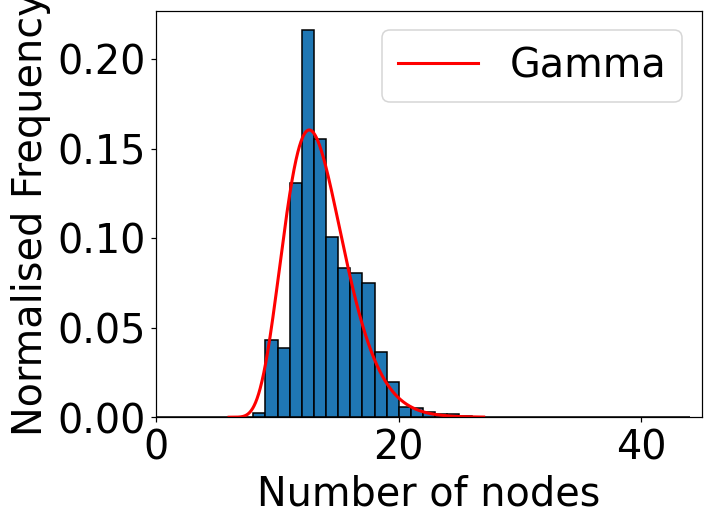}
        \caption{Node distribution.}
        \label{fig:accessn}
    \end{subfigure}
    \hfill
    \begin{subfigure}{0.49\columnwidth}
        \centering
        \includegraphics[width=\columnwidth]{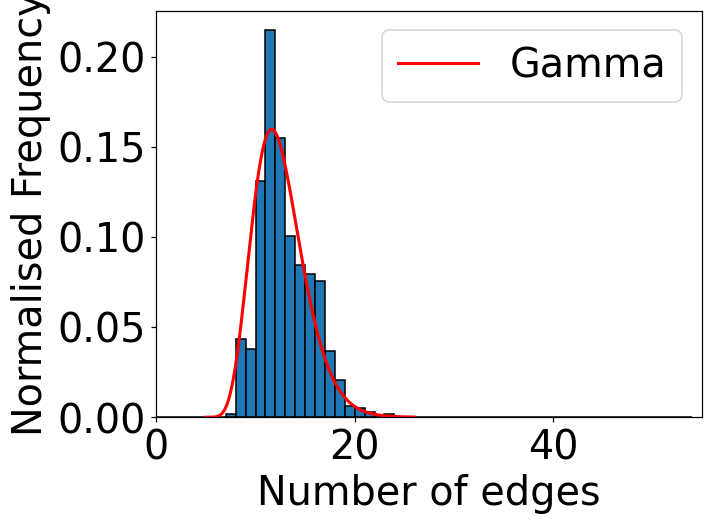}
        \caption{Edge distribution.}
        \label{fig:accesse}
    \end{subfigure}

    \caption{Histograms showing the node and edge distributions for accessibility graphs.}
    \label{fig:access_histograms}
\end{figure}

\begin{figure}
    \centering

    \begin{subfigure}{0.49\columnwidth}
        \centering
        \includegraphics[width=\linewidth]{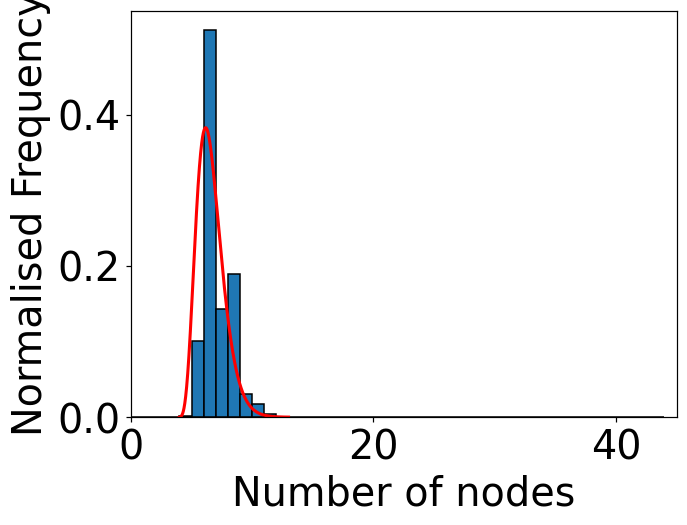}
        \caption{Node distribution.}
        \label{fig:adjn}
    \end{subfigure}
    \hfill
    \begin{subfigure}{0.49\columnwidth}
        \centering
        \includegraphics[width=\linewidth]{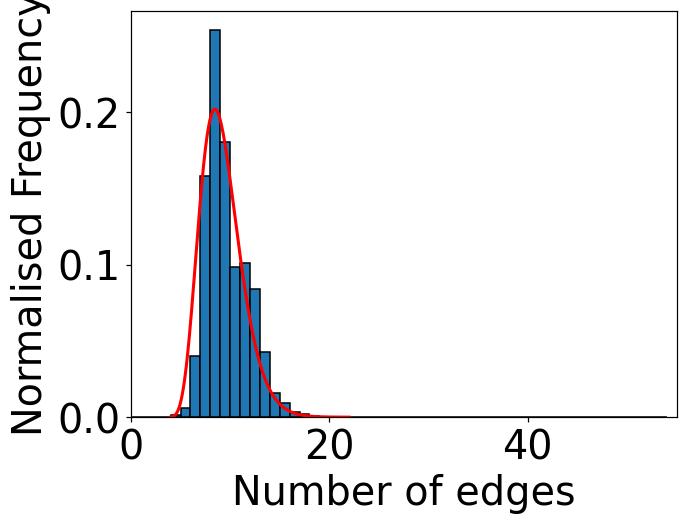}
        \caption{Edge distribution.}
        \label{fig:adje}
    \end{subfigure}

    \caption{Histograms showing the node and edge distributions for adjacency graphs.}
    \label{fig:adj_histograms}
\end{figure}

\begin{figure}
    \centering

    \begin{subfigure}{0.49\columnwidth}
        \centering
        \includegraphics[width=\linewidth]{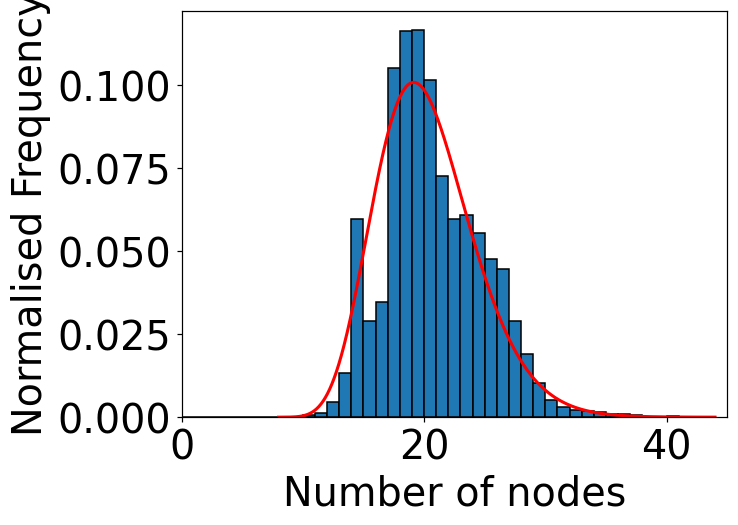}
        \caption{Node distribution.}
        \label{fig:fulln}
    \end{subfigure}
    \hfill
    \begin{subfigure}{0.49\columnwidth}
        \centering
        \includegraphics[width=\linewidth]{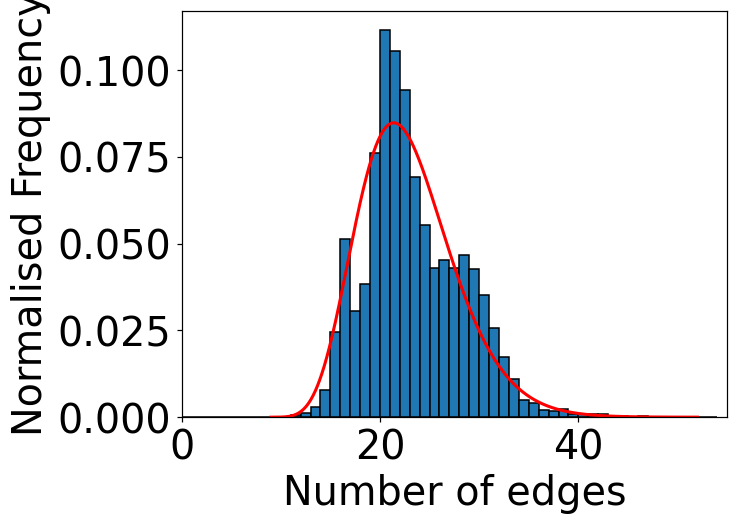}
        \caption{Edge distribution.}
        \label{fig:fulle}
    \end{subfigure}

    \caption{Histograms showing the node and edge distributions for full graphs.}
    \label{fig:full_histograms}
\end{figure}

We consider the distribution of the number of nodes and edges across our three graph types (See Figures \ref{fig:access_histograms}, \ref{fig:adj_histograms}, \ref{fig:full_histograms}). Adjacency graphs consistently have the least number of nodes and edges, with the lowest variance, followed by accessibility graphs, and then full graphs, which have the greatest number of nodes with the largest variance. This is expected, as adjacency graphs only contain nodes for each room, and doors are added for accessibility graphs, and then windows are added for full graphs, increasing both the number and variability of nodes and edges.

\begin{figure}
    \centering
    \includegraphics[width=0.5\linewidth]{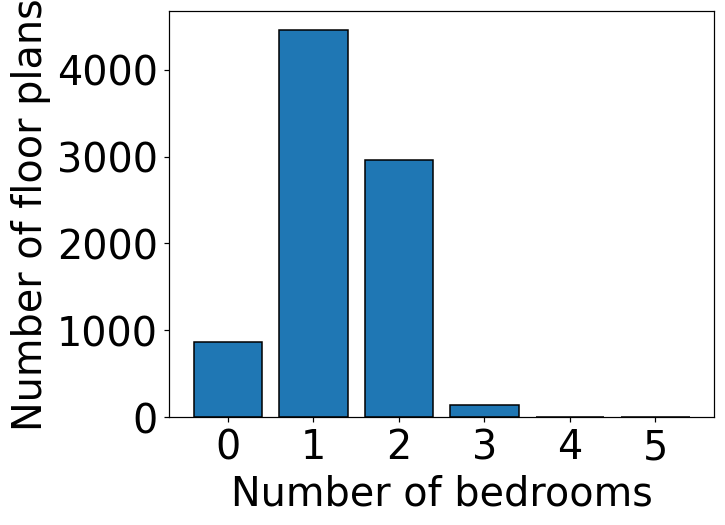}
    \caption{Frequency of different numbers of bedrooms in ModRes.}
    \label{fig:bedrooms}
\end{figure}

The distribution for the number of nodes and edges for each graph type can be approximated using a Gamma distribution. Each of these distributions has a skew of 0.6-0.8, which is indicative of a moderate right-skew (See Table 5 and Table 6). When compared to the distribution for number of bedrooms, we can see that this skew can be explained by the large number of two-bedroom houses within our dataset (See Figure \ref{fig:bedrooms}). While this may limit the applicability of models trained on this data to larger houses, it is reflective of real-life housing distributions. For example, $63.7 \%$ of houses in the USA are two- and three-bedroom houses \citep{acs2024}. Thus, the dataset captures a realistic, albeit unbalanced, representation of residential buildings.
\begin{table}
    \centering
    \begin{tabular}{|m{1.5cm}|m{1cm}|m{1cm}|m{1.5cm}|m{1cm}|} \hline
        Graph & Mean & Variance & Standard Deviation & Skew\\ \hline
        Full & 20.34 & 16.90 & 4.11 & 0.57\\ \hline
        Adjacency & 6.60 & 1.26 & 1.12 & 0.80\\ \hline
        Accessibility & 13.46 & 6.79 & 2.61 & 0.66\\ \hline
    \end{tabular}
    \caption{Variables for the gamma distribution approximation for the number of nodes in each type of graph.}
    \label{tab:gammanodes}
\end{table}
\begin{table}
    \centering
    \begin{tabular}{|m{1.5cm}|m{1cm}|m{1cm}|m{1.5cm}|m{1cm}|} \hline
        Graph & Mean & Variance & Standard Deviation & Skew\\ \hline
        Full & 23.06 & 24.68 & 4.97 & 0.63\\ \hline
        Adjacency & 9.19 & 4.29 & 2.07 & 0.65\\ \hline
        Accessibility & 12.52 & 6.89 & 2.62 & 0.65\\ \hline
    \end{tabular}
    \caption{Variables for the gamma distribution approximation for the number of edges in each type of graph.}
    \label{tab:gammaedges}
\end{table}

\begin{figure}
    \centering
    
    % Top row
    \begin{subfigure}[t]{0.49\columnwidth}
        \centering
        \includegraphics[width=\columnwidth]{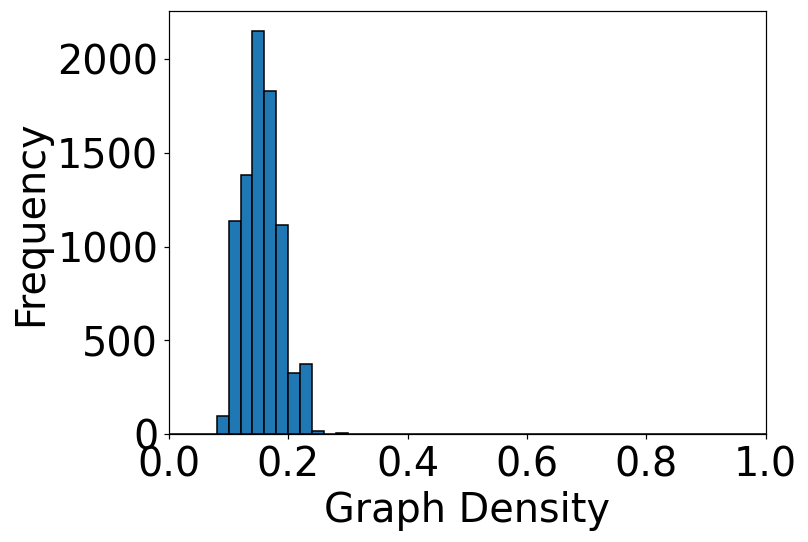}
        \caption{Accessibility Graphs.}
        \label{fig:accessd}
    \end{subfigure}
    \hfill
    \begin{subfigure}[t]{0.49\columnwidth}
        \centering
        \includegraphics[width=\columnwidth]{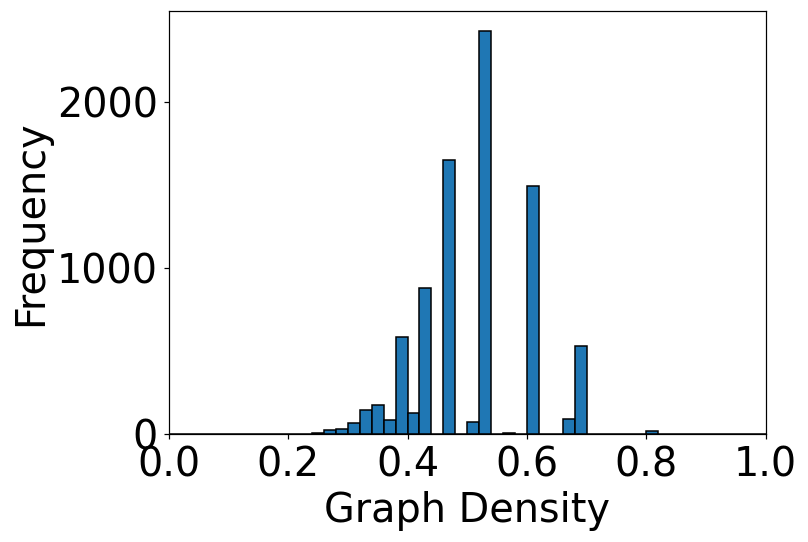}
        \caption{Adjacency Graphs}
        \label{fig:adjd}
    \end{subfigure}

    \vspace{0.5cm} % space between rows

    % Bottom row (centered)
    \begin{subfigure}[t]{0.49\columnwidth}
        \centering
        \includegraphics[width=0.9\columnwidth]{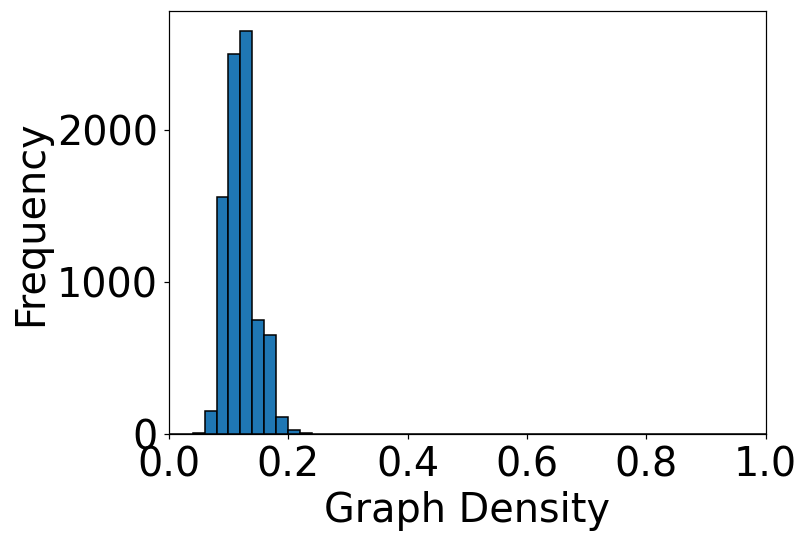}
        \caption{Full Graphs}
        \label{fig:fulld}
    \end{subfigure}

    \caption{Histograms to show the distribution of density of graphs across all three graph types.}
    \label{fig:densityhist}
\end{figure}

\begin{figure}
    \centering
    
    % Top row
    \begin{subfigure}[t]{0.49\columnwidth}
        \centering
        \includegraphics[width=\columnwidth]{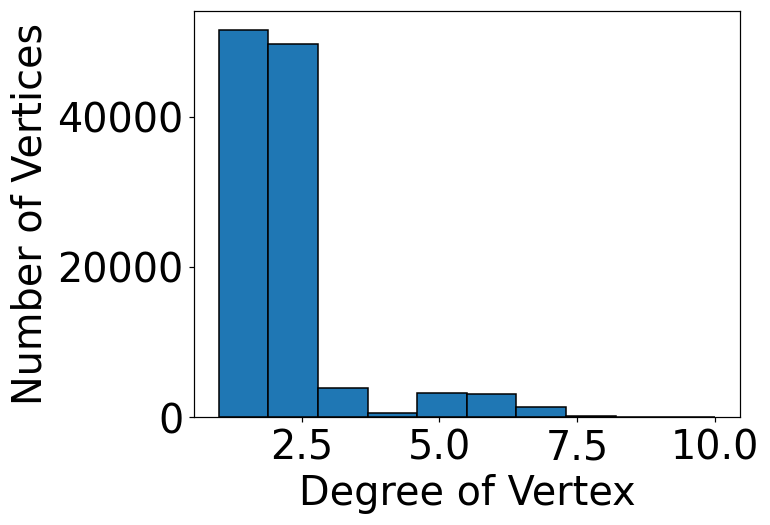}
        \caption{Accessibility Graphs}
        \label{fig:accessdeg}
    \end{subfigure}
    \hfill
    \begin{subfigure}[t]{0.49\columnwidth}
        \centering
        \includegraphics[width=\columnwidth]{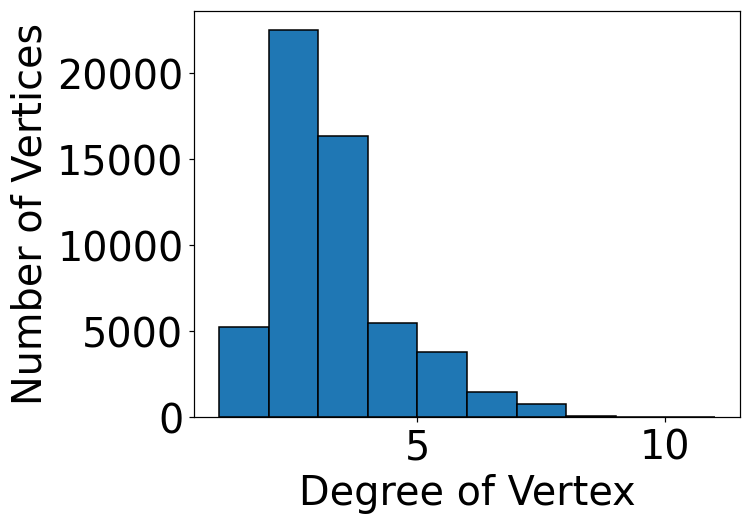}
        \caption{Adjacency Graphs}
        \label{fig:adjdeg}
    \end{subfigure}

    \vspace{0.5cm} % space between rows

    % Bottom row (centered)
    \begin{subfigure}[t]{0.49\columnwidth}
        \centering
        \includegraphics[width=\columnwidth]{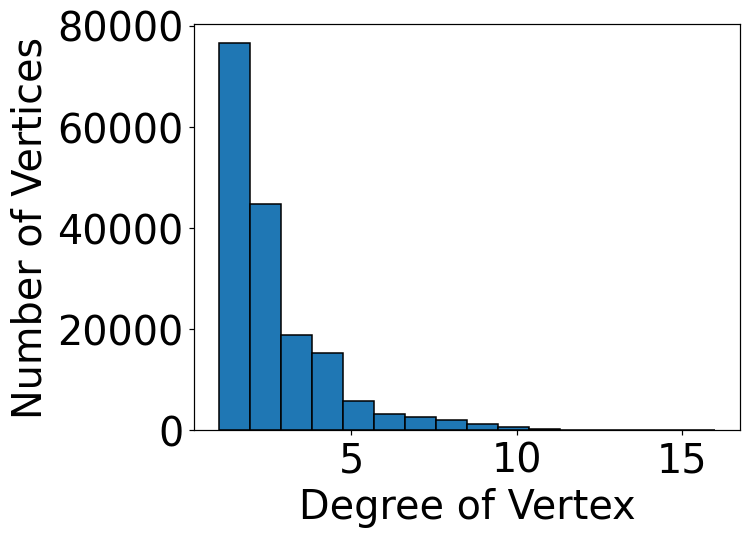}
        \caption{Full Graphs}
        \label{fig:fulldeg}
    \end{subfigure}

    \caption{Histograms to show the distribution of the degrees of nodes across all three graph types.}
    \label{fig:degrees}
\end{figure}

As well as size-based metrics, we also consider the structure of the floor plans. While graphs may have similar numbers of nodes and edges, they can vary greatly in how these nodes are connected to each other. Thus, we look at the degrees of each vertex within the graphs.

In Figure \ref{fig:degrees}, we see that the degrees of the vertices are skewed to the right, with a majority of vertices having low degrees. In the case of Accessibility and Full Graphs, this skewness can be explained through the inclusion of doors and windows, as vertices representing external doors and windows all have degree 1, and internal doors have degree 2. However, Adjacency graphs only contain nodes for rooms, so are more representative of how rooms are connected to each other.

In Figure \ref{fig:degrees}, we still see a bias towards low degree vertices, meaning most rooms are not connected to many other rooms. However, there are some very high degree vertices, up to 10, representing 'hubs' which are connected to a large number of rooms. The bias towards low degrees can again be explained by the number of two-bedroom floor plans within the dataset, as two bedroom floor plans often have much simpler structure than a floor plan with many more rooms. 

Next, we look at the density, $D$, of each graph. This is calculated as $$ D=\frac{2|E|}{|V|(|V|-1)},$$ where $|V|$ is the number of nodes in the graph, and $|E|$ is the number of edges. $D$ has a value between 0 and 1, with 1 being a dense graph, where each node is connected to all other nodes, and 0 being a graph with no edges.

\begin{figure}
    \centering
    \begin{subfigure}[t]{0.49\columnwidth}
        \centering
        \includegraphics[width=\columnwidth]{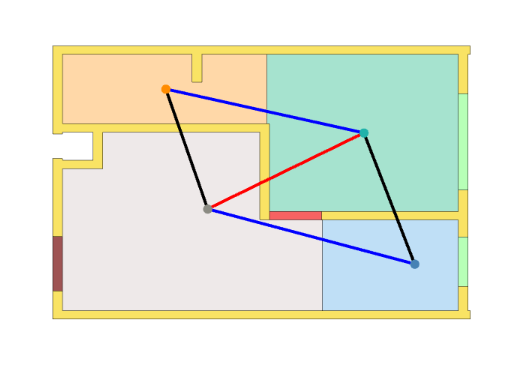}
        \caption{$D = 0.83$.}
        \label{fig:highdensity}
    \end{subfigure}
    \hfill
    \begin{subfigure}[t]{0.49\columnwidth}
        \centering
        \includegraphics[width=\columnwidth]{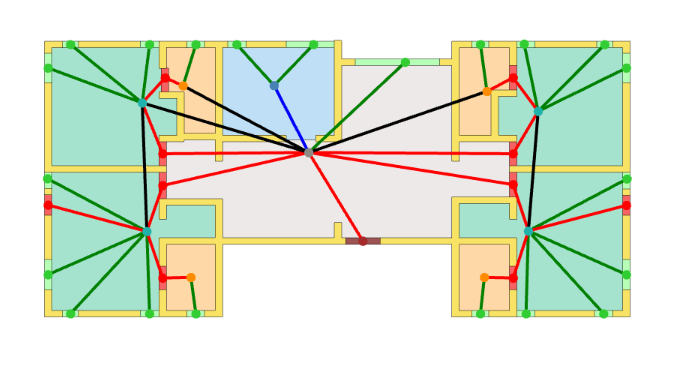}
        \caption{$D = 0.05$.}
        \label{fig:lowdensity}
    \end{subfigure}
    \caption{Examples of high and low density graphs from the Modified ResPlan dataset.}
    \label{fig:density}
\end{figure}

We observe in Figure \ref{fig:densityhist} that both Accessibility graphs and Full graphs have low density, whereas Adjacency graphs have higher density. Figure \ref{fig:density} shows examples of high and low density graphs. In Figure \ref{fig:highdensity} we have a very high density graph, as it is only one edge away from being a complete graph. Figure \ref{fig:lowdensity} shows a low density graph, as there are many windows that are only connected to one room. This is why all the full graphs are low density, as they include windows, which are all only connected to one node, i.e. the room to which they are attached.

\section{Use Test Cases} \label{energy}
ModRes can be used in a variety of applications in the Architecture, Engineering and Construction industry. Here, we outline some examples.

\subsection{Energy analysis}

To run energy analysis on a building, we use EnergyPlus \citep{energyplus}. EnergyPlus is a whole-building energy simulation program that can model and determine the energy consumption of a building. The energy consumption of a building is given as heating and cooling loads. The heating load of a building is the amount of energy required to heat a building to a comfortable temperature on the coldest day of the year. Conversely, the cooling load is the amount of energy required to cool a building to a comfortable temperature on the hottest day of the year. EnergyPlus incorporates a range of factors that can influence the energy consumption, such as weather conditions, building materials and insulation, and internal gains from occupants. 

One of the key features of EnergyPlus is its modular framework. EnergyPlus contains many individual modules that each model a specific aspect of the building's energy performance, such as heat transfer, HVAC components, or daylight. These modules are then controlled by a central simulation manager, which controls the interactions between all the individual simulations.

EnergyPlus calculates the heating and cooling loads by using a predictor-corrector approach to model the interaction between thermal conditions and HVAC systems. Heat balance equations are solved at sub-hourly time steps, to estimate the air temperature based on previous time steps. These predicted values are then used to simulate the response of the HVAC system. In the corrector step, temperature is then updated to include the HVAC system's response, subject to the limits of the system's capacity and capabilities. 

To use EnergyPlus to find heating and cooling loads for each of the floor plans within the dataset, we use the 3D representation of the building. Other inputs are also required. Firstly, we require a weather file. Since ResPlan was originally made in the United States, we choose a weather file from Columbus Ohio. 

We also require an OpenStudio file, which specifies the materials that the building is made from, and the material's U-values. A material's U-value is a measure of how quickly heat can travel through the material, with lower values indicating slower heat transmission, and, thus, higher insulation. We produce a residential template, which uses typical constructions and U-values for a residential building, based on US building regulations. This OpenStudio file is shared along all other data sources and code.

The outputs of the energy simulations are the heating and cooling loads. We normalise these to get values for heating and cooling load, per square metre, for each floor plan. Heating and cooling load have been used in previous research to train machine learning algorithms to predict the energy efficiency of buildings \citep{jabi}.
\begin{figure}
    \centering
    \begin{subfigure}[t]{0.49\columnwidth}
        \centering
        \includegraphics[width=\columnwidth]{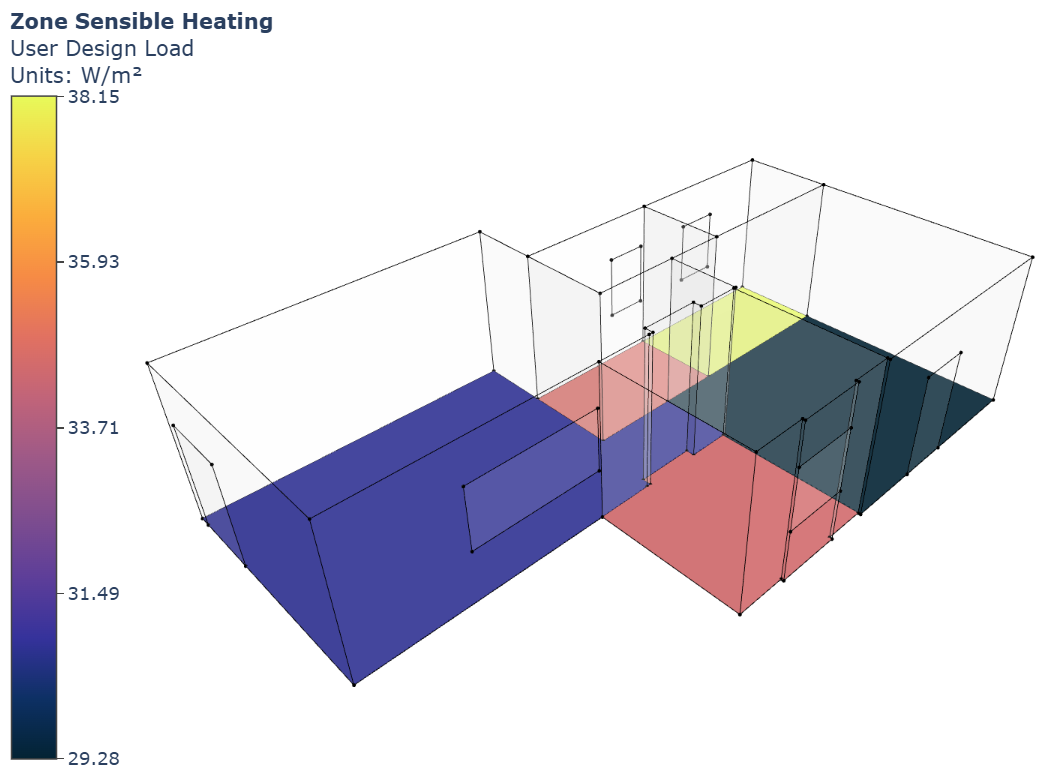}
        \caption{Heating load}
        \label{fig:heating}
    \end{subfigure}
    \hfill
    \begin{subfigure}[t]{0.49\columnwidth}
        \centering
        \includegraphics[width=\columnwidth]{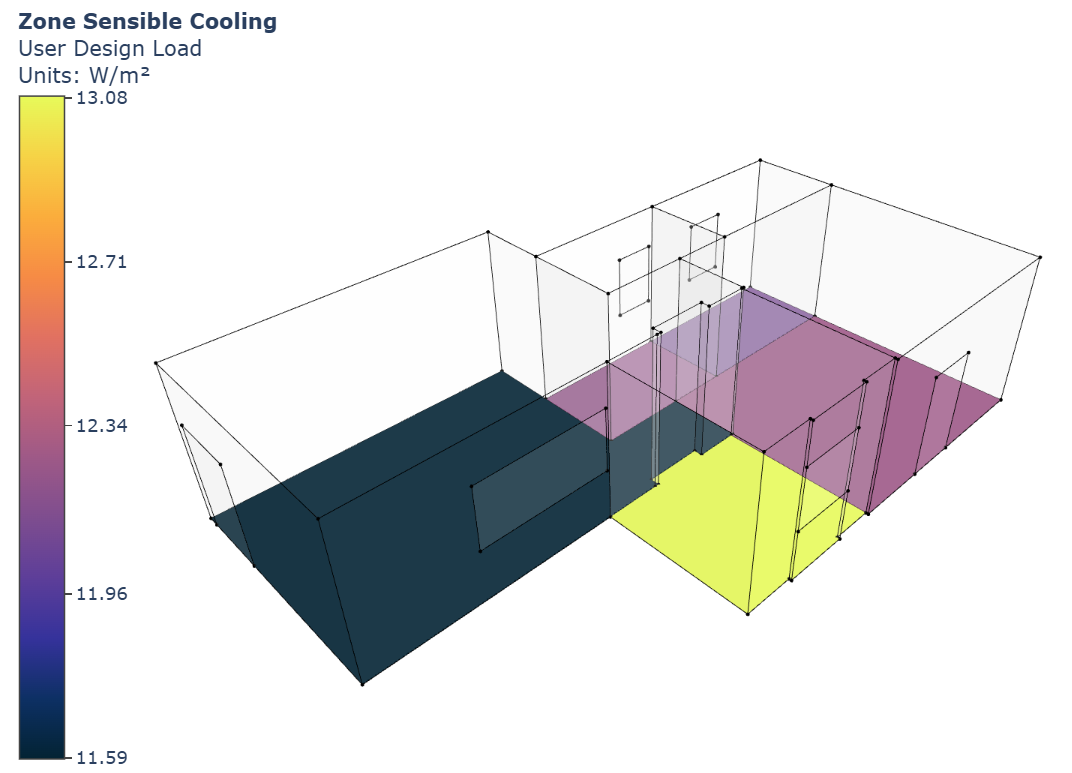}
        \caption{Cooling load}
        \label{fig:cooling}
    \end{subfigure}
    \caption{Heat maps showing the heating and cooling load per square metre for plan 67 in the Modified ResPlan dataset.}
    \label{fig:main}
\end{figure}

\subsection{Space Syntax}
\begin{figure}
    \centering
    \begin{subfigure}[t]{0.49\columnwidth}
        \centering
        \includegraphics[width=\columnwidth]{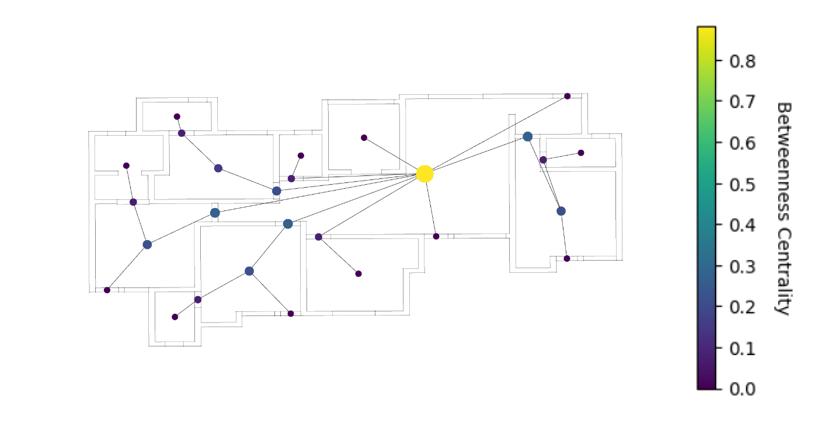}
        \caption{Betweenness Centrality of nodes.}
        \label{fig:betweennode}
    \end{subfigure}
    \hfill
    \begin{subfigure}[t]{0.49\columnwidth}
        \centering
        \includegraphics[width=\columnwidth]{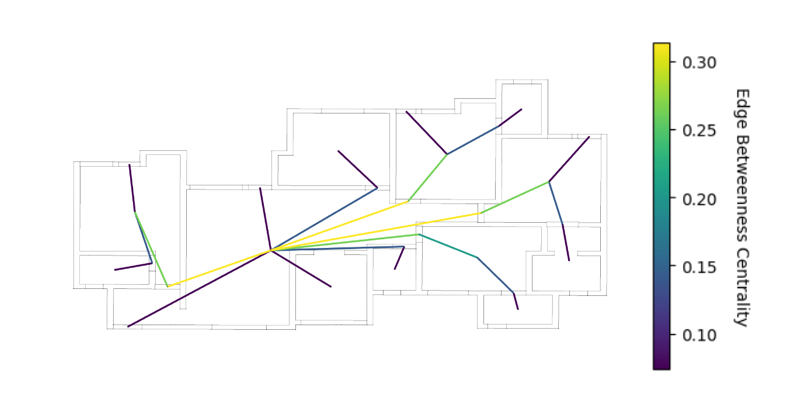}
        \caption{Betweenness Centrality of Edges.}
        \label{fig:betweenedge}
    \end{subfigure}
    \caption{Betweenness centrality for nodes and edges calculated on the accessibility graph for plan 8426.}
    \label{fig:between}
\end{figure}

Space syntax is a spatial analysis technique in which floor plans are represented as a network that captures spatial relationships between different areas of a building \citep{keles}. This technique can be used to analyse how people may move through a building. ModRes contains accessibility graphs for each floor plan which show how an individual can move through a building. Betweeness centrality can be calculated for these graphs to quantify the extent to which particular nodes or edges lie on the shortest path between other spaces. Using betweenness centrality, we can identify areas which are more connected to the rest of the house, and those which are more isolated. Consequently, nodes or edges with high betweenness centrality can be identified as important routes through the building (See Figure \ref{fig:between}).

\subsection{Visibility analysis}
\begin{figure}
    \centering
    \includegraphics[width=0.5\linewidth]{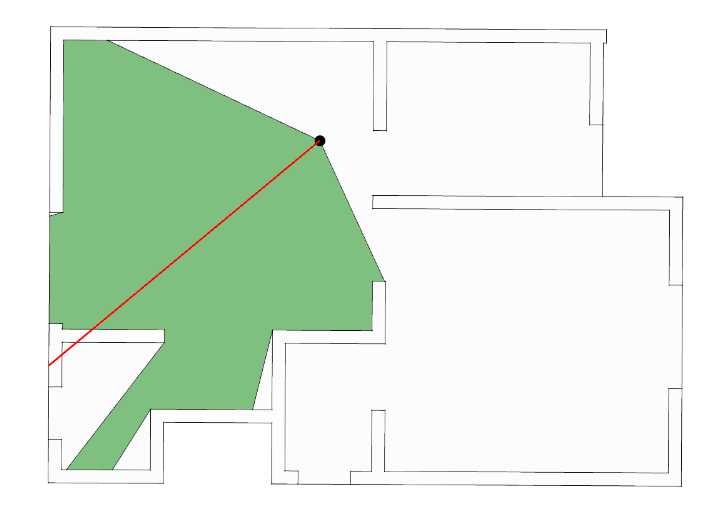}
    \caption{Visibility analysis on plan 7. The green area is visible from the given vertex, and the red line represents the direction of sight.}
    \label{fig:isovist}
\end{figure}

Using the 2D geometry of a floor plan in ModRes we can carry out visibility analysis. This will show us what area of the floor plan is visible from a given point (See Figure \ref{fig:isovist}). Such analysis has been used in architectural research to investigate properties such as visual privacy in a house \citep{isovist}. 

Together, these techniques demonstrate how the geometrical, topological and semantic data in ModRes can be used to derive higher level spatial information from a floor plan. The combination of these representations allows ModRes to act as a benchmark dataset for the development and evaluation of computational methods developed for the analysis of residential floor plans. In particular, the scale of the dataset enables such methods to be developed and evaluated across a wide range of diverse layouts, instead of on individual case studies. This scale provides a foundation for the development of automated applications such as compliance checking and design analysis, which can then be evaluated on new designs.

\section{Limitations} \label{lims}
While ModRes expands the capabilities of the original ResPlan dataset, and broadens the range of potential applications, there are still some key limitations that must be considered. 

\paragraph{Single floor.} Each floor plan within this dataset is a single-storey building. This limits the generalisability of any results obtained using this dataset to buildings with one floor. In reality, many residential buildings have more than one floor. Expanding the dataset to include multi-storey buildings, and their vertical connections through stairs, is one important avenue for future work.

\paragraph{3D assumptions.} ResPlan does not explicitly include 3D information, when converting each of the floor plans to 3D assumptions were made, such as ceiling height and roof geometry. These assumptions may not be representative of each individual building's properties, and may affect results that depend on spatial measurements, such as energy usage simulations. In the future, these measurements could also be included in the metadata of each floor plan, allowing for more accurate 3D representations.

\paragraph{Diversity of house types.} The dataset contains a large number of two-bedroom houses (See Figure \ref{fig:bedrooms}). While this is representative of real-life residences, it may limit the applicability of any machine learning algorithms trained on this data to larger homes. This limitation could be mitigated by stratified sampling, or by adding more houses with three or more bedrooms in the future. Furthermore, all of these houses are of the same typology; they all originate from North America.

\paragraph{Lack of furniture.} The dataset does not include any data on furniture or interior fixtures within the house, such as cabinets, tables, or appliances. This information can be important in applications such as indoor navigation and robotics, where furniture placement affects the accessible floor space within the building.

\paragraph{Graph Types.} While ModRes supplies three dual graphs for each floor plan, there are still more graphs that could be added. For example, medial axis graphs, in which nodes and edges correspond to skeletal structures of walls, could be added \citep{medial}. This addition would further broaden the scope of potential applications of the dataset.

\section{Conclusions and Future Work}

We have introduced ModRes, a cleaned and enriched extension of the ResPlan dataset for computational floor plan analysis. Starting from the original ResPlan data, we removed invalid, disconnected, unrealistic and duplicate floor plans, and corrected geometric inconsistencies that would otherwise limit the reliability of downstream analysis. The resulting dataset contains 8,431 valid and unique residential layouts, supplied with corrected two-dimensional geometry, image outputs, three-dimensional representations and three alternative dual graph representations. In doing so, ModRes reduces the preprocessing burden on users and provides a more robust basis for spatial analysis, graph machine learning and building-performance modelling.

The contribution of ModRes should be understood in the context of a wider need for reliable, interoperable and reusable floor plan datasets. Existing resources such as CubiCasa5K, RPLAN, MSD and ResPlan have enabled significant progress in automated floor plan understanding, but they also demonstrate that no single representation is sufficient for the full range of tasks now being asked of architectural data \citep{cubi,rplan,msd,resplan}. Rasterised images are useful for computer vision tasks, vectorised geometries support geometric and spatial analysis, graph representations support relational learning and spatial reasoning, and three-dimensional representations are required for many physically grounded simulation workflows. ModRes brings these representations together within a single cleaned dataset, allowing the same residential layouts to be studied from geometric, topological, visual and energy perspectives. This interoperability supports the development of automated workflows that can move from floor plan interpretation and spatial analysis to building performance assessment without requiring the dataset to be reconstructed for each task.

A central feature of the dataset is the provision of three dual graph types for each floor plan. These different graph structures are not interchangeable, since each encodes a different interpretation of the building. This is critical for Graph Machine Learning, where the choice of nodes, edges and features can strongly influence what a model is able to learn \citep{hu,kong,jabi}. By providing multiple graph abstractions in both CSV and JSON formats, ModRes allows users to select a representation that is appropriate to their task, and also provides a test bed for comparing how different graph constructions affect downstream performance.

The dataset nevertheless has limitations that should guide its future development. Future data sets should address these limitations by incorporating multi-storey layouts, richer semantic annotations, more detailed building-envelope metadata and more diverse housing stock.

ModRes provides a practical resource for data-driven research on residential floor plans. Its main value is not only that it corrects and extends ResPlan, but that it makes a single collection of floor plans available in several mutually consistent forms. This allows researchers to move more easily between spatial analysis, graph-based modelling and building simulations.

\section*{Data availability}
Dataset and associated code are available on GitHub at \url{https://github.com/emmashelley/ModRes}.

\section*{Funding acknowledgments}
This work was supported by the Engineering and Physical Sciences Research Council [grant number: EP/W524682/1].

\bibliographystyle{unsrtnat}

\bibliography{bibfile}

\end{document}